\documentclass{lcs2}

\usepackage{inconsolata}

\usepackage{graphicx}

\usepackage{amsmath,amsfonts,bm}

\def\eqref#1{equation~\ref{#1}}

\def\1{\bm{1}}

\DeclareMathAlphabet{\mathsfit}{\encodingdefault}{\sfdefault}{m}{sl}
\SetMathAlphabet{\mathsfit}{bold}{\encodingdefault}{\sfdefault}{bx}{n}

\PassOptionsToPackage{numbers, compress}{natbib}
\PassOptionsToPackage{dvipsnames, table}{xcolor}

\usepackage{hyperref}       
\usepackage{url}            
\usepackage{booktabs}       
\usepackage{amsmath}
\usepackage{amsfonts}       
\usepackage{nicefrac}       
\usepackage{microtype}      
\usepackage{xcolor}         

\usepackage{microtype}
\usepackage{graphicx}
\usepackage{booktabs} 
\usepackage{titlesec}
\usepackage{wrapfig}
\usepackage{hyperref}
\usepackage{enumitem}
\usepackage{minitoc}
\usepackage[ruled,vlined, linesnumbered]{algorithm2e}

\setlist[itemize]{noitemsep, topsep=0pt}

\usepackage{todonotes}
\usepackage{url}
\usepackage{subfig}
\usepackage{multirow}
\usepackage{array}
\usepackage{arydshln} 
\usepackage{amssymb}
\usepackage{tikz}
\usepackage{wrapfig}
\usepackage{array}
\usepackage{xcolor}
\usepackage{float}
\usepackage{sidecap}

\usepackage{amsmath, bm}
\usepackage{amsmath, bm}
\usepackage{amssymb}
\usepackage{mathtools}
\usepackage{amsthm}

\papergithub{https://github.com/parmanu-lcs2/pruning\_laws}  
\paperwebsite{https://parmanu.lcs2.in}  

\papertitle{Pruning Laws for Large Language Models} 

\papershortauthors{A. Sengupta and S. Chaudhary and T. Chakraborty} 
\paperauthors{%
  Ayan Sengupta \textsuperscript{\tiny 1}\equalcontrib \quad 
  Siddhant Chaudhary \textsuperscript{\tiny 1}\equalcontrib \quad
  Tanmoy Chakraborty \textsuperscript{\tiny {1,2}}%
} 

\paperaffil{\textsuperscript{1} Department of Electrical Engineering, Indian Institute of Technology Delhi \\ \textsuperscript{2} Yardi School of Artificial Intelligence, Indian Institute of Technology Delhi \\
* Equal contribution} 

\paperemails{{\ttfamily{ayan.sengupta@ee.iitd.ac.in, urssidd@gmail.com, tanchak@iitd.ac.in}}} 

\paperkeywords{Downscaling laws, efficient language models, model pruning} 

\paperabstract{ 
  Scaling up model parameters and training data consistently improves the performance of large language models (LLMs), but at the cost of rapidly growing memory and compute requirements, which makes deployment on resource-limited hardware infeasible. \textit{Model pruning}, a widely used compression technique, reduces inference costs by removing redundant parameters. However, its impact on downstream performance remains unpredictable and is typically assessed only through costly empirical sweeps. To address this gap, we introduce \emph{pruning laws} -- simple and interpretable scaling relations that connect a pruned LLM's post-pruning performance to its unpruned performance and pruning ratio. Across ten LLMs (1.3B–30B parameters){\color{black}, a 20B mixture-of-experts model}, three pruning strategies (unstructured, width, and depth), and eight diverse tasks, we show that pruning laws achieve strong predictive accuracy (average extrapolation error $<7\%$), reliably quantify performance degradation, and identify critical pruning thresholds beyond which recovery is infeasible. Moreover, we demonstrate that {\color{black}the functional form transfers across dense and mixture-of-experts architectures, pruning methods, and unseen models in zero-shot and one-shot setups, with task- and method-specific coefficients that vary in interpretable ways}. These results provide both researchers and practitioners with a principled framework to select pruning strategies, estimate safe pruning ratios without exhaustive tuning, and deploy LLMs efficiently under real-world compute and latency constraints.
} 

\appendixtocon 
\appendixtocname{Structure of the Appendix}   
\begin{document}

\makelabtitle


\section{Introduction}
\label{sec:intro}

\begin{figure*}[!htb]
\centering
\subfloat[Downstream model performance at different pruning ratios]{\includegraphics[width=0.85\linewidth]{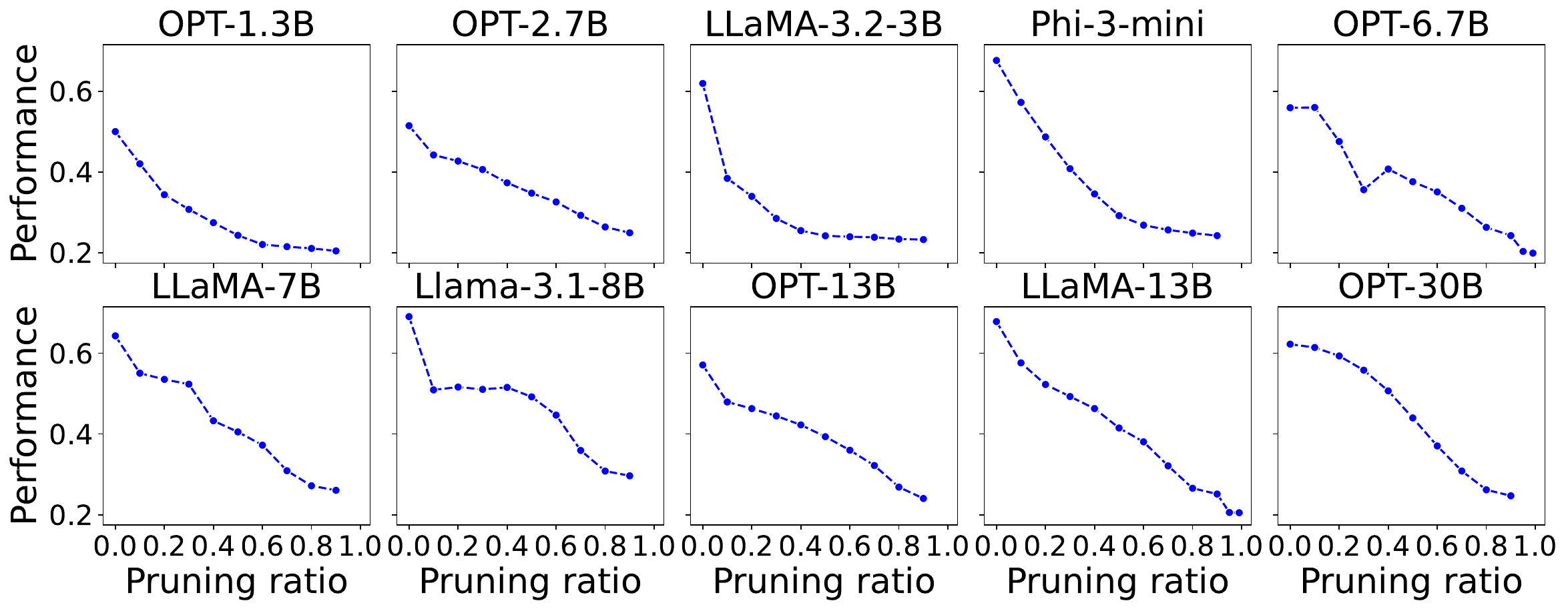}\label{fig:model_wise_avg_performance}}
\quad 
\subfloat[Inference speedup of pruned LLMs]{\includegraphics[width=0.85\linewidth]{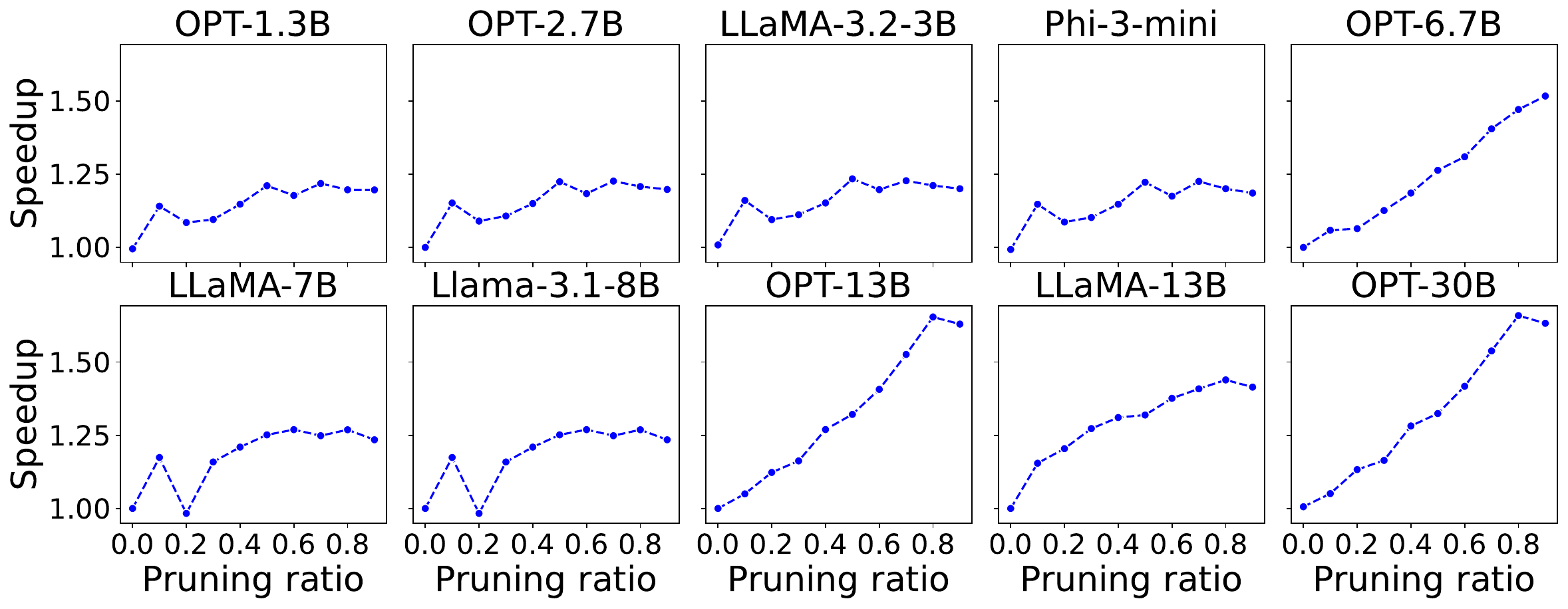}\label{fig:model_wise_runtime}}
    \caption{\textbf{A.} Average downstream performance of different LLMs at varying pruning ratios. The different performance degradation patterns emphasize the need for a systematic evaluation of pruning effects on different LLMs of different scale and architectural family. Full result highlighted in Figure~\ref{fig:model_wise_avg_performance_full} of Appendix~\ref{appx:results}. \textbf{B.} Inference speedup of different models at varying pruning ratios on an autoregressive language modeling benchmark.}
\end{figure*}

The rapid evolution of Large Language Models (LLMs) has been guided largely by \textit{neural scaling laws}~\citep{kaplan_scaling_2020, hoffmann_training_2022}, which predict that increasing model size and data volume consistently improves performance. However, this pursuit of scale has led to prohibitive computational costs, making deployment on resource-limited hardware increasingly infeasible~\citep{faiz_llmcarbon:_2024}. To address this, \textit{model pruning} has emerged as a critical technique for compressing LLMs. Whether through unstructured weight removal~\citep{frantar2023sparsegptmassivelanguagemodels} or structured approaches like depth and width pruning~\citep{yang2024lacolargelanguagemodel, ashkboos2024slicegptcompresslargelanguage}, pruning aims to eliminate redundancy without sacrificing capability. Yet, despite its widespread adoption, pruning remains an empirical art; there is currently no systematic framework to predict how a specific pruning strategy will impact downstream performance across different models and tasks.

To bridge this gap, we introduce \textbf{Pruning Laws} - a principled analytical framework that models post-pruning performance as a function of the pruning ratio. By formalizing these relationships, we move beyond costly trial-and-error sweeps to answer four fundamental questions:
\begin{itemize}[leftmargin=*]
\setlength{\itemsep}{0pt}
    \item \textbf{RQ1.} How does pruning affect downstream performance across task categories?
    \item \textbf{RQ2.} How do pruning strategies trade off between efficiency and accuracy?
    \item \textbf{RQ3.} Can pruning laws accurately predict performance for unseen models?
    \item \textbf{RQ4.} How can pruning laws guide principled pruning decisions?
\end{itemize}

Our study fits these laws across a diverse landscape of ten LLMs (including OPT, LLaMA-2/3, and Phi-3), three distinct pruning strategies, and eight downstream tasks. We define the pruning law using a power function: $\mathcal{L} = \mathcal{L}_0 P_0 (1-r)^\alpha$. This connects the pruned model's performance $\mathcal{L}$ to its unpruned baseline $\mathcal{L}_0$ and the retention ratio $(1-r)$, with $r$ being the pruning ratio. Our key empirical observations are summarized below.

\paragraph{RQ1. Impact on downstream performance.} 
We find that performance degradation follows a predictable power law, but the sensitivity varies drastically by task. As shown in Figure~\ref{fig:model_wise_avg_performance}, \textit{reasoning tasks} are remarkably resilient, characterized by a low decay exponent ($\alpha \approx 0.22$) that allows models to retain strong capabilities even at $50\%$ pruning. In stark contrast, \textit{QA tasks} are highly fragile ($\alpha \approx 2.42$), suffering sharp collapses at moderate compression ratios. Language modeling falls in between ($\alpha \approx 0.73$). Furthermore, we show that recovery fine-tuning (RFT) significantly mitigates this decay, reducing the sensitivity of language tasks from $0.73 \to 0.47$ and lowering test errors across the board.

\paragraph{RQ2. Computational benefits.} 
Our analysis quantifies the "no free lunch" trade-off in pruning. \textit{Depth pruning} offers the highest reward, achieving super-linear speedups ($>5\times$ at $90\%$ pruning) reflected by high scaling coefficients ($\alpha \approx 0.72$), though it suffers from higher variance in reliability. \textit{Unstructured pruning} is the safest for accuracy but offers negligible speedups ($\sim 1.2\times$) due to its flat runtime scaling ($\alpha \approx 0.01$). \textit{Width pruning} occupies the middle ground, providing stable, predictable speedups of $\sim 1.4\times$ alongside moderate accuracy retention.

\paragraph{RQ3. Generalization to newer models.} 
A critical feature of pruning laws is that the {\color{black}same functional form is reusable across setups}. When applied zero-shot to unseen architectures like LLaMA-3.1 and Phi-3, our laws predict performance with low error ($0.09$--$0.12$). For out-of-distribution pruning methods, a simple one-shot calibration, requiring just a single evaluation, further refines these predictions (e.g., reducing SlimGPT prediction error from $0.13 \to 0.05$). {\color{black}The form also holds on a mixture-of-experts model, GPT-OSS-20B, where removing a routed block restructures the active-parameter graph rather than narrowing a dense stack, yielding coefficients ($\alpha = 0.48$, $P_0 = 0.89$) inside the range spanned by the dense models. What transfers is the form; the coefficients themselves remain task- and method-specific.}

\paragraph{RQ4. Practical implications.} 
These laws provide actionable guidelines for practitioners. Instead of guessing, one can use our coefficients to plan safe pruning thresholds -- aggressive ratios ($60$--$70\%$) for reasoning-heavy applications, but conservative ones ($<40\%$) for QA. This framework transforms pruning from an ad-hoc procedure into a predictable engineering discipline.\footnote{Additional data, source code and dataset have been open-sourced at \url{https://github.com/parmanu-lcs2/pruning_laws}.} 

\section{Related Work}
\label{sec:related}

\paragraph{Model Pruning for Efficiency.} 
Large Language Models have demonstrated remarkable capabilities, yet their deployment is often bottlenecked by immense computational requirements. Model pruning addresses this by reducing parameter counts, generally falling into two categories. \textit{Unstructured pruning} removes individual weights~\citep{frantar2023sparsegptmassivelanguagemodels,sun2023simple}; while effective at preserving accuracy, it often requires specialized hardware to realize speedups. In contrast, \textit{structured pruning} removes entire components such as layers or channels~\citep{ashkboos2024slicegptcompresslargelanguage,yang2024lacolargelanguagemodel}, offering direct latency gains on standard hardware. Recent work has also explored calibration-free methods~\citep{sengupta2025pruneoncedesigningcalibrationfree} to simplify the compression pipeline. However, despite the variety of methods, practitioners currently lack a unified framework to predict how these strategies will impact performance without running exhaustive benchmarks.

\paragraph{Neural Scaling Laws.} 
The study of scaling laws has traditionally focused on pre-training. Foundational works like \textit{Kaplan scaling laws}~\citep{kaplan_scaling_2020} and \textit{Chinchilla scaling laws}~\citep{hoffmann_training_2022} established power-law relationships linking test loss to model size and data volume. Later studies have nuanced these findings, with \citet{caballero_broken_2023} noting that monotonic laws may fail to capture emergent behaviors in Transformers. 
More recently, research has pivoted toward \textit{parameter-efficient scaling}. \citet{busbridge2025distillation} introduced distillation scaling laws to optimize compute allocation between teacher and student models. Similarly, \citet{kumar_scaling_2024} proposed precision-aware scaling laws, demonstrating that low-precision training can improve scaling efficiency. \citet{chen2024scaling} examined scaling in the context of recovery fine-tuning for structured pruning. While these works offer valuable insights into training and recovery dynamics, they do not address the fundamental degradation characteristics of the pruning process itself.

\paragraph{Positioning Pruning Laws.} 
Our work introduces a distinct framework that complements existing compression theories. Unlike distillation scaling laws, which depend on specific teacher-student pairings and extensive retraining, or quantization scaling laws~\citep{cao2024scaling}, which are often tightly coupled to hardware backends, our \textbf{Pruning Laws} are architecture-agnostic and require no retraining. By modeling performance as a direct function of the pruning ratio, our framework allows practitioners to identify \textit{critical pruning thresholds} -- the point beyond which recovery becomes infeasible. This capability generalizes across diverse model families and pruning methods (including unseen ones), making it a robust tool for resource-aware deployment planning.
\section{Methodology}
\label{sec:method}


\subsection{Parametrization of the LLM pruning law}

With pruning laws, we propose a series of analytical methods for estimating the performance of LLM post-pruning on a variety of downstream tasks.  For all downstream tasks, we assume that the performance of a model on a task is captured by a metric, which is bounded (e.g., model accuracy); in other words, higher the performance of a model, the better it is.  Building on prior studies \citep{kaplan_scaling_2020, hoffmann_training_2022} that establish scaling laws for LLM pre-training, we formulate a relationship between the performance of a pruned model and its corresponding base model through a law defined by two key parameters: the performance of the base model on a task (denoted by \(\mathcal{L}_0\)) and the pruning ratio used to prune the model (denoted by \(r\in (0, 1)\)); we denote the relationship by \(\mathcal{L} := \mathcal{L}(\mathcal{L}_0, r)\), where \(\mathcal{L}\) represents the performance of the pruned model.  Our proposed pruning law can be used to determine the optimal value of \(r\) needed to obtain a well-performing pruned model, maximizing the performance retainment post-pruning.

The functional form of our parametrization is described by the following equation:
\begin{equation}
    \label{eq:main_fit}
    \mathcal{L}(\mathcal{L}_0, r) \;=\; \mathcal{L}_0\,P_0\,(1 - r)^\alpha
\end{equation}
where \(\alpha\) and \(P_0\) are real numbers.  The exponent \(\alpha\) controls how quickly performance decays as the pruning ratio \(r\) increases.  The coefficient \(P_0\) is a \textit{bias term} that encodes the boundary conditions of the law: although \((1-r)\) tends to 1 as \(r\) approaches 0, the performance of the pruned model at extremely small pruning ratios may not exactly match the base model’s performance.  Factors such as evaluation noise, subtle architecture-dependent effects and differences among pruning methods can introduce a systematic offset at \(r\to 0\).  Consequently, \(P_0\) captures these offsets and depends on the particular model, pruning method and task being considered.


\textbf{Levels of fit.}  In our analysis, we fit the pruning law at several levels of aggregation:

\begin{enumerate}[leftmargin=*]
\setlength{\itemsep}{0pt}
    \item \textbf{Task level:} we fit a single \((\alpha, P_0)\) pair for each task, pooling data across all models and pruning methods.  In this case, \(P_0\) reflects the average boundary-condition bias for that task.
    
    \item \textbf{Method–task level:} we fit the law separately for each pruning method and task across all models.  Here \(P_0\) absorbs both task-specific and method-specific biases (e.g., unstructured vs. structured pruning may yield different offsets near \(r=0\)).
    
    \item \textbf{Model–task level:} we fit the law for each model and task, pooling across methods.  \(P_0\) then captures architecture-specific bias for that task, indicating how resilient a given model is on that task when lightly pruned.
    
\end{enumerate}

This choice of functional form is motivated by the following \textit{feasibility conditions} for a pruning law:
\begin{itemize}[leftmargin=*]
\setlength{\itemsep}{0pt}
\item We formulate the pruning law as a power law with respect to the retention ratio.  Unlike some pre-training scaling laws, we require the law to be \textit{scale invariant} with respect to $(1 - r)$, i.e.,\ \(\mathcal{L}(\mathcal{L}_0, r)\) must be a homogeneous function of $(1 - r)$.  Such a form allows us to derive optimal decision regions for choosing \(r\) given a target performance drop.
\item The post-pruning performance should decrease as the pruning ratio \(r\) increases, i.e.,\ \(\frac{\partial \mathcal{L}}{\partial r} < 0\).  In Equation~\ref{eq:main_fit}, this behavior is captured by requiring \(\alpha > 0\).
\item The functional form should capture the \textit{iterative nature} of many pruning methods: pruning a model by \(r_1\) followed by further pruning by \(r_2\) should yield a performance proportional to a single pruning with overall pruning ratio \(1 - (1-r_1)(1-r_2)\) \citep{sengupta2025pruneoncedesigningcalibrationfree, yang2024lacolargelanguagemodel}.
\end{itemize}

Here, we show that our proposed functional form in Equation ~\ref{eq:main_fit} satisfies the last feasibility condition for iterative pruning methods. So, let $r_1\in(0, 1)$ and $r_2\in(0, 1)$ be two pruning ratios. Suppose a model is first pruned at ratio $r_1$, followed by a pruning by the same method at ratio $r_2$. It is easy to see that the overall pruning ratio (\textit{i.e.}, a pruning ratio achieving the total pruning in one step) is $r_1 + r_2 - r_1r_2$. Then, the performance obtained by the multi-step pruning is
\begin{align*}
    \mathcal{L}_\text{mutlistep} &= \mathcal{L}(\mathcal{L}(\mathcal{L}_0, r_1), r_2)\\
    &= \mathcal{L}(P_0\mathcal{L}_0(1 - r_1)^\alpha, r_2)\\
    &= P_0^2\mathcal{L}_0(1 - r_1)^\alpha(1 - r_2)^\alpha\\
    &= P_0\mathcal{L}(\mathcal{L}_0, r_1 + r_2 - r_1r_2)\\
    &= P_0\mathcal{L}_\text{singlestep}
\end{align*}
Hence, our functional form attains proportional performance between single-step and multi-step pruning for iterative methods. Through extensive experiments (see Section~\ref{sec:results}), we find that the proposed form in Equation ~\ref{eq:main_fit} satisfies remaining of the feasibility conditions.


\subsection{Fitting the pruning law: Ordinary least squares for linear regression}

Taking logarithms on both sides of Equation~\ref{eq:main_fit} gives:
\begin{equation}
    \small
    \log \mathcal{L}
    \;=\;
    \log \mathcal{L}_0
    \;+\;
    \alpha\,\log (1 - r)
    \;+\;
    \log P_0,
\end{equation}
Therefore, fitting the pruning law reduces to a linear regression in log space.  In this formulation, \(\alpha\) is the slope (multiplying \(\log(1-r)\)) and \(\log P_0\) is the intercept, capturing the boundary-condition bias discussed above.  The regression is performed on the transformed variables \(r' := 1 - r\) with \(\log P_0\) as the intercept.  To learn \(\alpha\) and \(P_0\), we use the standard ordinary least squares (OLS) method \citep{Zdaniuk2014}, assuming additive Gaussian noise.  The regression model can thus be stated as:
\begin{equation}
    \log \mathcal{L}
    \;=\;
    \log \mathcal{L}_0
    \;+\;
    \alpha\,\log(1 - r)
    \;+\;
    \log P_0
    \;+\;
    \epsilon_{\text{noise}},
\end{equation}
where \(\epsilon_{\text{noise}} \sim \mathcal{N}(0,1)\).\footnote{A similar implementation with robust estimation Huber loss \citep{Huber1992} is also performed, but due to very marginal differences omitted from this paper.}  

\section{Experimental Setup}
\label{sec:experimental_setup}




To develop the proposed pruning laws, we compress and evaluate a range of LLMs, including OPT~\citep{zhang2022optopenpretrainedtransformer} at 1.3B, 2.7B, 6.7B, 13B and 30B parameter scales, and LLaMA-2~\citep{touvron2023llama} at 7B and 13B. Moreover, for evaluating the universality of the pruning laws on more recent architectures, we consider LLaMA-3.1-8B, LLaMA-3.2-3B~\citep{dubey2024llama} and Phi-3-mini-4K-Instruct~\citep{abdin2024phi} models. All pre-trained model checkpoints are obtained from Hugging Face~\footnote{\url{https://huggingface.co/models}}. We experiment with three pruning methods: (1) \textbf{SparseGPT}~\citep{frantar2023sparsegptmassivelanguagemodels} for unstructured weight pruning, (2) \textbf{LaCo}~\citep{yang2024lacolargelanguagemodel} for structured depth pruning (layer collapse), and (3) \textbf{SliceGPT}~\citep{ashkboos2024slicegptcompresslargelanguage} for structured width pruning. Additional testing is conducted with two further depth pruning methods -- \textbf{SlimGPT}~\citep{ling2024slimgpt} and \textbf{LLM Pruner}~\citep{ma2023llmpruner} and two width pruning methods including \textbf{SVD-LLM}~\citep{wang2024svd} and \textbf{PruneNet}~\citep{sengupta2025pruneoncedesigningcalibrationfree}. {\color{black}To test the laws beyond dense decoder-only architectures, we further run a full depth pruning sweep on \textbf{GPT-OSS-20B}~\citep{openai2025gptoss}, an open-weights mixture-of-experts (MoE) model, using \textbf{ShortGPT}~\citep{men2024shortgptlayerslargelanguage}, which ranks layers by a Block-Influence score.} For each method, we apply a comprehensive range of sparsity levels with pruning ratios of $\{10\%, 20\%, \dots, 90\%\}$. We evaluate the pruned models on eight downstream tasks using the LM Evaluation Harness~\citep{eval-harness}~\footnote{Task descriptions are provided in Appendix~\ref{appx:dataset}.}, performing all evaluations in a zero-shot setting.
The tasks span three categories:

\begin{itemize}[leftmargin=*]
\setlength{\itemsep}{0pt}
    \item \textbf{Reasoning}: PIQA~\citep{bisk2020piqa}, WinoGrande~\citep{sakaguchi2021winogrande}, HellaSwag~\citep{zellers2019hellaswag}, ARC-e and ARC-c~\citep{clark2018think}.
    \item \textbf{Question-answering}: CoQA~\citep{reddy2019coqa}
    \item \textbf{Language modeling}: WikiText~\citep{merity2016pointer} and LAMBADA~\citep{paperno2016lambada}.
\end{itemize}

For classification-style tasks, including reasoning and QA, we report \textit{accuracy}. For language modeling tasks, we compute \textit{perplexity}, which is unbounded $(1,\infty)$; thus, we report the inverse log-perplexity (\text{i.e.}, $1/\log(\text{ppl})$) to map the metric to a $(0,1)$ range for consistency with other evaluations. In addition to these individual tasks, we calculate the average task performance, denoted as `\textbf{Average}'. 

\paragraph{Evaluating pruning laws.} For evaluating the fitted pruning laws on in-distribution data points (\text{e.g.,} same model/pruning method used in training and testing), we use \textit{rolling-style} test data, where for each pruning ratio $r \in [20\%, 90\%]$, we train the pruning laws on pruning ratios $\{10\%, \cdots, r\%\}$ and test the pruning laws on $\{r\%+10\%, \cdots 90\%\}$. Finally, we calculate the average root mean square error (RMSE) between the predicted and ground truth over all the test datasets. For evaluating the extrapolation capability of the pruning laws in out-of-distribution setup, we adopt two extrapolation evaluation strategies -- \textit{zero-shot extrapolation}, where we use the fitted parametric functions to test on unseen data, and \textit{one-shot extrapolation}, where we use the $\alpha$ coefficient from the fitted parametric function, but the bias term $P_0$ is re-estimated from pruning the model on a single pruning ratio. Precisely, let $\alpha$ and $P_0$ denote the fitted coefficients. For a given pruning ratio $r$, we prune the LLM, obtain its performance $\mathcal{L}_{r}$, and re-estimate the bias term $\hat{P}_0 = \frac{\mathcal{L}_r}{\mathcal{L}_0 \times (1-r)^\alpha}$. While zero-shot extrapolation setup is intended to verify the universality of the pruning laws, one-shot extrapolation is particularly useful for verifying the flexibility of pruning laws for testing out-of-distribution models or pruning strategies, when a single point enumeration is practically possible.
\section{Experimental Results}
\label{sec:results}

\begin{table}[!t]
  \centering
  \scalebox{0.76}{
    \begin{tabular}{lrrrr}
    \cline{1-5}
    \textbf{Model} & \multicolumn{1}{l}{\textbf{Reasoning}} & \multicolumn{1}{l}{\textbf{QA}} & \multicolumn{1}{l}{\textbf{Language}} & \multicolumn{1}{l}{\textbf{Average}} \\
    \cline{1-5}
    OPT-1.3B & 0.53 & 0.45 & 0.44 & 0.50 \\
    OPT-2.7B & 0.52  & 0.51  & 0.49  & 0.51 \\
    LLaMA-3.2-3B & 0.66 & 0.64 &  0.52 & 0.62 \\
    Phi-3-mini & 0.74  & 0.68  & 0.53  & 0.68 \\
    OPT-6.7B & 0.57  & 0.55  & 0.54  & 0.56 \\
    LLaMA-7B & 0.64  & 0.64  & 0.64  & 0.64 \\
    LLaMA-3.1-8B & 0.69  & 0.68  & 0.69  & 0.69 \\
    OPT-13B & 0.58  & 0.56  & 0.56  & 0.57 \\
    LLaMA-13B & 0.67  & 0.66  & 0.69  & 0.68 \\
    OPT-30B & 0.64 & 0.56 & 0.60 & 0.62 \\
    \bottomrule
    \end{tabular}%
    }
  \caption{Results obtained by the unpruned models on different tasks.}
  \label{tab:base_model_results}%
\end{table}%


\begin{figure*}[t]
    \centering
    \includegraphics[width=\linewidth]{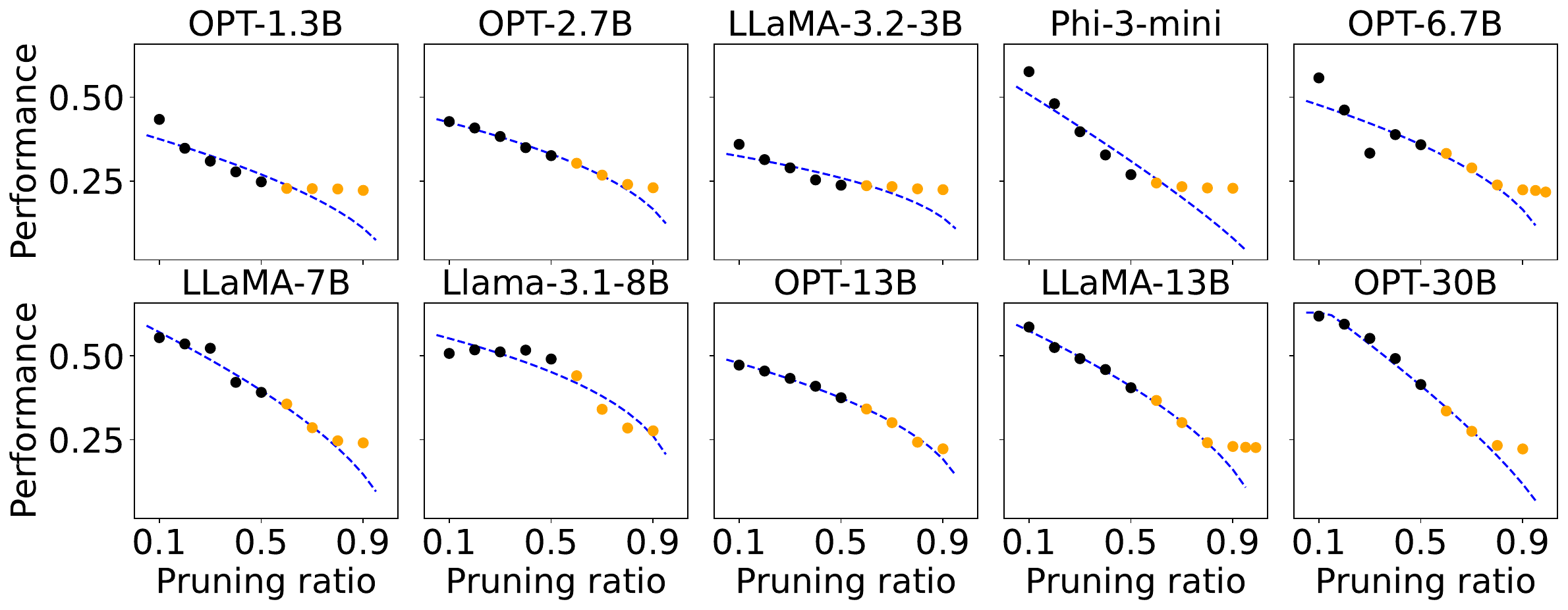}
    \caption{Fitted pruning laws for average downstream performance of pruned LLMs. Task-specific visualizations reported in Figure~\ref{fig:model_wise_avg_scaling_full} of Appendix~\ref{appx:results}. \textbf{Black} and \color{orange}orange \color{black}points highlight the training and testing data points, respectively, with \color{black}black \color{black} line indicating the scores predicted by our pruning laws.}
    \label{fig:model_wise_avg_scaling}
\end{figure*}

\subsection{Unpruned LLM Performance}

Table~\ref{tab:base_model_results} reports the accuracy of ten unpruned models across reasoning, QA, and language modeling tasks. As expected, performance generally improves with scale: OPT-2.7B achieves an average of $0.51$, whereas OPT-13B and LLaMA-13B improve to $0.57$ and $0.68$, respectively. The LLaMA series consistently outperforms OPT at similar parameter scales, underscoring architectural and training advantages. Notably, Phi-3-Mini-4K-Instruct matches the average of LLaMA-13B ($0.68$) despite being smaller in size, and LLaMA-3.1-8B surpasses both ($0.69$). These results confirm that newer families, even at comparable or smaller parameter counts, can rival or outperform older architectures. These baseline numbers provide the $\mathcal{L}_0$ values against which pruning performance is evaluated.

\subsection{Pruned LLM Results}


\paragraph{Impact of pruning on model performance.} 
Pruning exerts a clear and systematic influence on downstream performance, with degradation patterns varying significantly across model scales and architectures. As illustrated in Figure~\ref{fig:model_wise_avg_performance}, larger models such as LLaMA-13B and OPT-30B exhibit notable resilience, maintaining competitive accuracy from high baselines (see Table~\ref{tab:base_model_results}) up to 30-40\% pruning. In contrast, smaller but highly efficient models like LLaMA-3.2-3B and Phi-3-mini degrade much more rapidly. Despite LLaMA-3.2-3B achieving a high initial average score of 0.62, Figure~\ref{fig:model_wise_avg_performance} reveals a sharp performance drop even at low pruning ratios ($<20\%$), confirming that while larger models offer greater tolerance due to parameter redundancy~\cite{frantar2023sparsegptmassivelanguagemodels}, highly optimized efficient architectures possess significantly less redundant capacity.

Regarding task sensitivity, reasoning capabilities tend to be more robust compared to QA or language modeling. Larger models like LLaMA-7B manage to retain relatively strong reasoning scores even at 50\% pruning, whereas QA and language modeling capabilities erode faster. However, regardless of model size, performance consistently collapses across all architectures once pruning ratios exceed 70-80\%, with average scores approaching random baselines. These observations suggest that reasoning-oriented tasks degrade more gracefully than linguistic knowledge tasks, implying that pruning disproportionately impacts the retrieval of specific linguistic information over logical problem-solving abilities.

\paragraph{Pruning laws for downstream performance.} 
Table~\ref{tab:all_scaling} demonstrates pruning laws that characterize how downstream performance scales with pruning ratios. Across tasks (cf. Table~\ref{tab:task_level}), the exponent $\alpha$ is highest for QA ($\alpha \approx 2.42$) and smallest for reasoning ($\alpha \approx 0.22$). This indicates that QA accuracy decays sharply with pruning, whereas reasoning remains relatively stable even under high compression. Language tasks fall in between ($\alpha \approx 0.73$), showing gradual degradation.

\begin{table}[!htb] 
\centering
\subfloat[Task-level\label{tab:task_level}]{
\scalebox{0.75}{
    \begin{tabular}{lccrrr}
    \toprule
    \bf Task  & $\alpha$ & $P_0$ & \bf $R^2$ & \bf F & \bf Error \\
    \midrule
    \bf QA    & 2.42 $\pm$ 0.1 & 1.45 $\pm$ 0.1 & 0.89  & 348.5 & 0.04 \\
    \bf Reasoning & 0.22 $\pm$ 0.0 & 0.95 $\pm$ 0.0 & 0.79  & 170.8 & 0.06 \\
    \bf Language & 0.73 $\pm$ 0.0 & 0.76 $\pm$ 0.0 & 0.96  & 947.9 & 0.03 \\
    \bf Average & 0.35 $\pm$ 0.0 & 0.85 $\pm$ 0.0 & 0.88  & 316.4 & 0.05 \\
    \bottomrule
    \end{tabular}%
 }
}
\quad
\subfloat[Method-task level \label{tab:method_level}]{
\scalebox{0.75}{
    \begin{tabular}{lccrrr}
    \toprule
    \textbf{Method} & \textbf{$\alpha$} & $P_0$ & \multicolumn{1}{l}{\textbf{$R^2$}} & \textbf{F} & \multicolumn{1}{l}{\textbf{Error}} \\
    \midrule
    Unstr & 0.47 $\pm$ 0.0 & 1.22 $\pm$ 0.0 & 0.90  & 370.2 & 0.15 \\
     Depth & 0.13 $\pm$ 0.0 & 0.54 $\pm$ 0.0 & 0.20  & 12.2 & 0.10 \\
    Width & 0.39 $\pm$ 0.0 & 0.83 $\pm$ 0.0 & 0.70  & 103.1 & 0.10 \\
    \bottomrule
    \end{tabular}%
    }
}
\quad
\subfloat[Model-task level\label{tab:model_level}]{
\scalebox{0.7}{
    \begin{tabular}{lccrrr}
      \toprule
    \bf Model & $\mathbf{\alpha}$ & $P_0$ & \bf $R^2$ & \bf F & \bf Error \\
    \midrule
    OPT-1.3B & 0.26 $\pm$ 0.1 & 0.70 $\pm$ 0.1 & 0.57  & 11.5 & 0.10 \\
    OPT-2.7B & 0.3 $\pm$ 0.0 & 0.83 $\pm$ 0.0 & 0.92  & 87.7 & 0.05 \\
    LLaMA-3.2-3B & 0.18 $\pm$ 0.0 & 0.48 $\pm$ 0.1 & 0.56  & 11.3 & 0.08 \\
    Phi-3-mini & 0.4 $\pm$ 0.1 & 0.64 $\pm$ 0.1 & 0.62  & 14.0 & 0.12 \\
    OPT-6.7B & 0.19 $\pm$ 0.0 & 0.74 $\pm$ 0.1 & 0.65  & 19.7 & 0.11 \\
    LLaMA-7B & 0.43 $\pm$ 0.0 & 0.85 $\pm$ 0.1 & 0.89  & 66.8 & 0.06 \\
    LLaMA-3.1-8B & 0.35 $\pm$ 0.0 & 0.81 $\pm$ 0.0 & 0.88  & 59.8 & 0.09 \\
    OPT-13B & 0.37 $\pm$ 0.0 & 0.86 $\pm$ 0.0 & 0.97  & 235.2 & 0.02 \\
    LLaMA-13B & 0.23 $\pm$ 0.0 & 0.69 $\pm$ 0.1 & 0.70   & 24.1 & 0.11 \\
    OPT-30B & 0.53 $\pm$ 0.1 & 0.99 $\pm$ 0.1 & 0.90   & 70.0 & 0.07 \\
    
    \bottomrule
    \end{tabular}%
 }
}

\caption{Coefficients of the (a) task-level, (b) method-task-level, and (c) model-task level pruning laws of the form $\mathcal{L} = \mathcal{L}_{0}P_{0}(1-r)^{\alpha}$. We report the coefficients estimated with ordinary least square (OLS) fits, along with the standard errors. To evaluate the goodness-of-fit, we calculate adjusted $R^2$, F statistic. We calculate the Error as the average root mean square error on the test data. Pruning laws with recovery fine-tuned LLMs are reported in Table~\ref{tab:all_scaling_rft} of Appendix~\ref{appx:results}. We further highlight the results with other functional forms in Table~\ref{tab:other_forms}, justifying the power-law functional form defined in Eq.~\ref{eq:main_fit}.}
\label{tab:all_scaling}
\end{table}

The bias term $P_0$, which modulates retained performance, is close to one for reasoning tasks ($P_0 \approx 0.95$), reflecting robustness. In contrast, it is substantially larger for QA ($P_0 \approx 1.45$), indicating disproportionately large performance drops. The deviation of $P_0$ from unity arises as a fitting artifact when a power law models ``cliff-like'' degradation curves: fragile tasks often maintain stability at low pruning ratios before collapsing sharply. To capture this steep degradation, the regression extrapolates a theoretical intercept ($P_0\mathcal{L}_0$) that exceeds the true baseline. Thus, $P_0 \gg 1$ signals a phase-transition-like behavior rather than super-compressibility. We also observe deviations at very small pruning ratios ($r < 10\%$), where empirical performance remains near $\mathcal{L}_0$ while the power law predicts slightly higher values. This indicates that the pruning law primarily characterizes the degradation regime rather than the stable low-sparsity regime.

\begin{figure*}[!htb] 
    \centering
    \includegraphics[width=\linewidth]{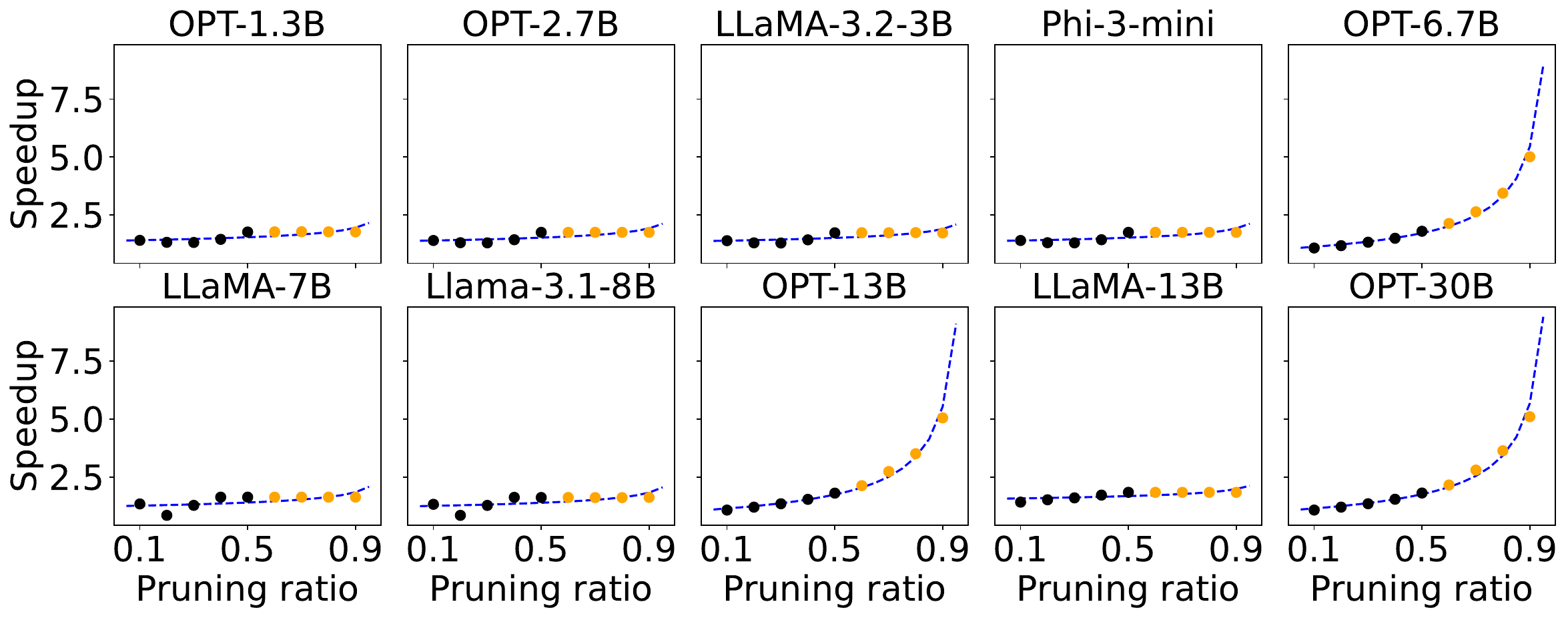}
    \caption{Fitted pruning laws for inference speedup of pruned LLMs for depth pruning. We report the remaining visualizations for other pruning methods in Figure~\ref{fig:model_wise_avg_scaling_runtime_full} of Appendix~\ref{appx:results}.}
\label{fig:model_wise_avg_scaling_runtime}
\end{figure*}

\paragraph{Dependence of model architecture on effectiveness of pruning.} 
Across model families (cf. Table~\ref{tab:model_level}), the pruning law maintains strong predictive accuracy, though sensitivity varies with architecture and scale. Larger models such as LLaMA-13B and OPT-30B exhibit stable bias terms ($P_0 \approx 0.69$--$0.99$), indicating strong performance retention at the onset of pruning. In contrast, smaller efficient models like LLaMA-3.2-3B show lower bias terms ($P_0 \approx 0.48$), reflecting early performance collapse at low pruning ratios ($<20\%$) before entering a smoother decay regime. While larger OPT models exhibit higher $\alpha$ values ($\approx 0.53$), this reflects decay from a higher baseline rather than increased fragility. The goodness-of-fit also correlates with scale; larger models achieve higher adjusted $R^2$ scores (e.g., $0.97$ for OPT-13B), confirming that the pruning law reliably captures architecture-specific behaviors. Overall, larger models offer stronger buffers against pruning by avoiding the catastrophic early drops observed in smaller architectures.

\paragraph{Does the pruning method matter?} 
Method-level coefficients reported in Table~\ref{tab:method_level} reveal clear differences among pruning strategies. Unstructured pruning shows the strongest adherence to the pruning law, with adjusted $R^2 = 0.90$ and a large $F$-statistic of 370.20. This strong fit likely arises from the fine-grained nature of weight removal, which produces a smooth reduction in model capacity that aligns well with power-law scaling. Depth pruning performs poorly in comparison, with adjusted $R^2 = 0.20$, a weak $F$-statistic of 12.21, and a low bias term ($P_0 \approx 0.54$). This behavior stems from the coarse-grained removal of entire layers, which introduces high variance depending on which layer is removed. Although the fit is weaker, empirical comparisons of functional forms (Table~\ref{tab:other_forms}) confirm that the power law still yields lower test error (0.05) than linear or polynomial alternatives (0.06--0.10). Width pruning lies between the two, with moderate $\alpha$ ($\approx 0.39$), stable $P_0$ ($\approx 0.83$), and adjusted $R^2 = 0.70$. These results indicate that the pruning law generalizes across strategies while distinguishing their behavior.

{\color{black}The weak depth pruning fit is, however, a property of the \textit{pruning operator} rather than of the framework. LaCo collapses layers according to a merge schedule that is discontinuous in $r$: the merge plan changes abruptly as $r$ crosses block-boundary thresholds, so the realized degradation does not interpolate smoothly between adjacent ratios and a smooth power law absorbs the jaggedness as residual variance. Consistent with this reading, the one depth-pruning fit we obtain with a smoother operator -- ShortGPT~\citep{men2024shortgptlayerslargelanguage}, which ranks layers by a continuous Block-Influence score -- reaches an average-task adjusted $R^2$ of $0.65$ (Table~\ref{tab:oss_results}), against $0.20$ for LaCo. The two are not matched in scope -- the ShortGPT fit covers one model, the LaCo row pools ten -- so this is indicative rather than a controlled ablation. It nonetheless suggests that the fit quality of the law inherits the smoothness of the operator, which is a useful diagnostic when selecting a pruner.}

\begin{figure*}[!htb]
    \centering
\includegraphics[width=\linewidth]{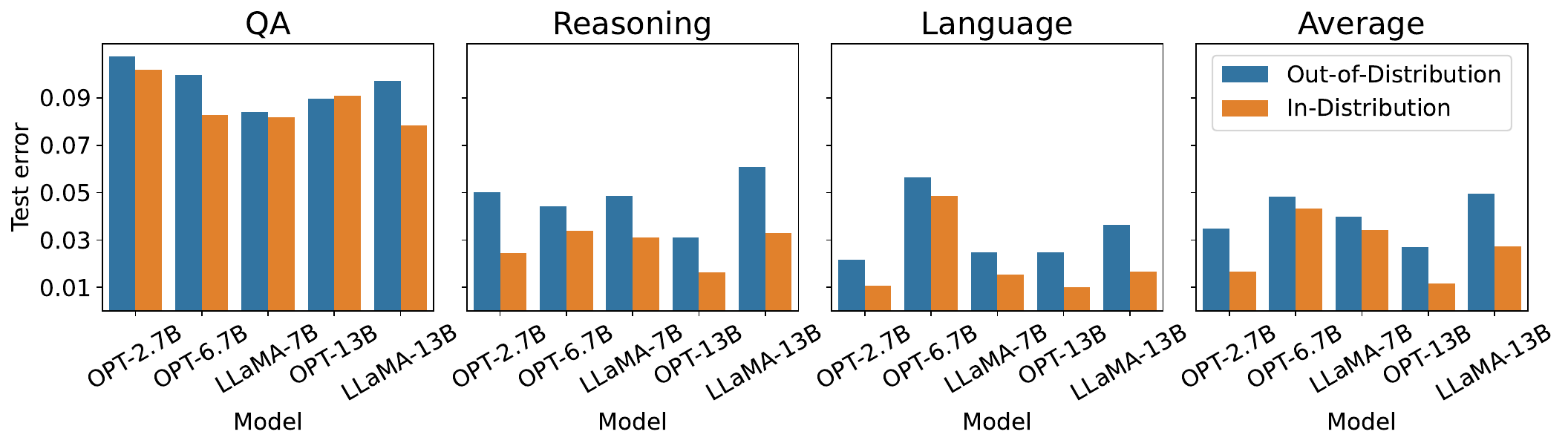}
    \caption{In-distribution (ID) and out-of-distribution (OOD) test error for different pruned LLMs. In the ID setup a model (\text{e.g.,} LLaMA-7B) is tested while a different model from the same family (\text{e.g.,} LLaMA-13B) is used in training. In the OOD setup, training and testing are performed on different model families.}
    \label{fig:model_wise_extrapolation}
\end{figure*}

\paragraph{Influence of model pruning on inference speed.} 
Pruning substantially improves inference speedup, though gains depend on pruning ratio and strategy (cf. Figure~\ref{fig:model_wise_avg_scaling_runtime} and Table~\ref{tab:runtime} of Appendix~\ref{appx:results}). Depth pruning delivers the largest benefits, with super-linear speedups at higher pruning ratios. This behavior is reflected in relatively high $\alpha$ values (e.g., $0.72 \pm 0.03$ for OPT-13B and $0.73 \pm 0.03$ for OPT-30B). In contrast, unstructured pruning yields smaller but stable improvements, consistent with its near-zero $\alpha \approx 0.01$ (e.g., $0.01 \pm 0.0$ for LLaMA-7B and OPT-6.7B). Width pruning falls between the two, with $\alpha \approx 0.14$--$0.15$ for larger models (e.g., $0.15 \pm 0.04$ for LLaMA-13B). Goodness-of-fit metrics reinforce these trends: unstructured pruning shows strong adjusted $R^2$ values ($0.94$--$0.97$) and minimal test error ($0$--$0.01$), whereas depth pruning exhibits weaker fits and higher variance in test error ($0.03$--$0.34$). Overall, pruning laws confirm that depth pruning can yield dramatic inference acceleration, albeit with less stability, while unstructured and width pruning provide steadier improvements.

{\color{black}These wall-clock gains are hardware-conditional. Depth and width pruning produce dense smaller sub-networks and realize their gains directly on standard accelerators. Unstructured pruning does not: fine-grained sparsity is not exploited by conventional dense kernels, and gains require N:M structured-sparse support~\citep{mishra2021nmsparsity}, so its near-flat $\alpha \approx 0.01$ reflects the backend we measure on rather than the sparsity itself. The two families of laws should therefore be read differently -- the performance law of Eq.~\ref{eq:main_fit} is hardware-independent, whereas the speedup law is conditional on the deployment kernel.}

\subsection{Universality of Pruning Laws}

Our results indicate that the same functional form $\mathcal{L}_{1}=\mathcal{L}_{0}\,(1-r)^{\alpha}P_{0}$ transfers across \emph{models}, \emph{architectures}, and \emph{pruning methods} with moderate testing error. Figure~\ref{fig:model_wise_extrapolation} clearly illustrates the consistently low in-distribution (ID) (when a model from same family is used in fitting the functions) test errors, as compared to out-of-distribution (OOD) (when none of the same-family models used in fitting) test errors. Particularly, for reasoning and language modeling tasks with larger models (13B sized), having same family models in the training data (in-distribution) reduces the test error by $50\%$. 

\begin{table}
\centering
\scalebox{1}{
    \begin{tabular}{clcc}
    \toprule
    \multicolumn{1}{l}{\textbf{Method Type}} & \textbf{Pruner} & \multicolumn{1}{l}{\textbf{0-shot error}} & \multicolumn{1}{c}{\textbf{1-shot error}} \\
    \midrule
    \multirow{2}[0]{*}{Depth} & LLM Pruner & 0.10   & 0.06 \\
          & SlimGPT & 0.13  & 0.05 \\
    \cdashline{1-4}
    \multirow{2}[0]{*}{Width} & SVD-LLM & 0.13  & 0.16 \\
          & PruneNet & 0.04  & 0.03 \\
    \bottomrule
    \end{tabular}%
 }
\caption{OOD extrapolation results. }
\label{tab:extrapolate_method}
\end{table}

We further test universality by applying our fitted pruning laws to completely new pruning methods not used during training (Table~\ref{tab:extrapolate_method}). We obtain the performances of LLaMA-1-7B model~\citep{touvron2023llamaopenefficientfoundation} on reasoning tasks for depth pruning methods LLM Pruner~\citep{ma2023llmpruner} and SlimGPT~\citep{ling2024slimgpt} and two width pruning methods -- SVD-LLM~\citep{wang2024svd} and PruneNet~\citep{sengupta2025pruneoncedesigningcalibrationfree}, all numbers obtained from the corresponding papers. In the zero-shot setup, extrapolation errors remain small, ranging from $0.04$ for PruneNet to $0.13$ for SlimGPT and SVD-LLM. In the one-shot setup, where only $\hat{P_0}$ is recalibrated, errors decrease for most methods (e.g., LLM Pruner: $0.10 \rightarrow 0.06$, SlimGPT: $0.13 \rightarrow 0.05$, PruneNet: $0.04 \rightarrow 0.03$). Overall, these results reinforce the universality of pruning laws: they transfer across unseen pruning algorithms in zero-shot mode, while one-shot calibration may improve performance when the method aligns well with the retained elasticity.

{\color{black}\paragraph{Transfer to a mixture-of-experts model.} All results so far concern dense decoder-only LLMs. To test whether the law survives a change of architecture, we fit it on GPT-OSS-20B, in which each block routes tokens to a subset of experts, so that removing a block restructures the active-parameter graph rather than narrowing a dense stack. Equation~\ref{eq:main_fit} applies without modification, yielding $\alpha = 0.48 \pm 0.13$ and $P_0 = 0.89 \pm 0.11$ on the average task -- both inside the range spanned by the dense models -- and the task ordering carries over (reasoning $0.30$, language $0.98$, QA $0.99$; Table~\ref{tab:oss_results} of Appendix~\ref{appx:results}). The sweep also separates two QA styles the main study could not: CoQA reproduces the cliff-like collapse discussed above ($\alpha = 3.53$, $P_0 \gg 1$), whereas CommonsenseQA decays gradually ($\alpha = 0.52$, $P_0 = 0.54$). The form covers both, with coefficients differing by an order of magnitude.}

Practically, this means practitioners can \emph{predict} pruned performance for new model families and pruning strategies with zero data, and, if desired, achieve better estimation with a single-point calibration of $P_{0}$. {\color{black}We stress that what transfers is the \textit{functional form}, not a single set of coefficients: across the architectures, methods and tasks we study, $\mathcal{L} = \mathcal{L}_0 P_0 (1-r)^\alpha$ remains the right two-parameter description, while $(\alpha, P_0)$ themselves remain task- and method-specific and vary in interpretable ways.}

\subsection{Practical Implications of Pruning Laws} 

Rather than relying on ad-hoc trial and error, our pruning laws provide a principled roadmap for model compression. Practitioners \emph{seeking speed can turn to depth pruning}, those \emph{prioritizing accuracy to unstructured pruning}, and those \emph{balancing both to width pruning}. When \emph{shifting to new models or pruning methods, zero-shot reuse of our coefficients enables immediate deployment}, while \emph{one-shot calibration offers sharper estimates if a single evaluation is possible}. By aligning pruning choices with task sensitivity, \emph{aggressive ratios for reasoning}, \emph{conservative ones for QA and language modeling}, practitioners can achieve reliable efficiency gains without sacrificing performance. We provide more detailed practical guidelines in Appendix~\ref{appx:guidelines} for the readers to better synthesize our results for practical real-life deployment situations.

\section{Conclusion}


In this paper, we introduced pruning laws for LLMs that characterize how downstream performance scales with pruning ratio across tasks, architectures, and pruning strategies. Our analysis reveals systematic relationships between sparsity and performance, enabling predictive modeling of the accuracy--efficiency trade-off during model compression. Empirically, pruning sensitivity increases with model scale, while recovery tuning can mitigate performance degradation even with limited data. Among the evaluated methods, depth pruning yields the largest inference speedups, whereas unstructured pruning methods provide stronger robustness for downstream performance. Overall, our work establishes a quantitative foundation for understanding pruning behavior in LLMs and motivates future research on adaptive-hybrid compression strategies that balance computational efficiency with performance stability.


\section*{Limitations}

While our proposed pruning laws provide a robust framework for predicting the performance of pruned LLMs, we acknowledge certain limitations that bound the scope of our findings. First, our empirical validation is conducted on models ranging from 1.3B to 30B parameters{\color{black}, together with a single 20B mixture-of-experts model}. While we observe consistent scaling behaviors across this range, confirming these laws on significantly larger foundation models (e.g., 70B+ {\color{black}dense or MoE} architectures) remains a subject for future study due to the high computational cost of such validations{\color{black}: a sweep at our resolution ($9$ ratios $\times$ multiple methods $\times$ $8$ tasks) on a 70B+ model is beyond the compute budget available to us}. 
Furthermore, regarding inference acceleration, our study highlights that unstructured pruning offers negligible speedups on standard hardware without specialized kernels. While the pruning laws accurately predict the \textit{potential} theoretical efficiency, the realization of these gains in practice is heavily dependent on the underlying hardware and software stack. 

{\color{black}\paragraph{When pruning laws fail.} Our results also delineate the boundary conditions of the framework, which we consolidate here as four failure modes together with the mitigation we would recommend in each case.

\begin{enumerate}[leftmargin=*] \setlength{\itemsep}{0pt}
    \item \textbf{Discontinuous pruning operators.} When the pruning schedule jumps as $r$ crosses structural thresholds, as in LaCo's layer-merging plan, degradation does not interpolate smoothly between adjacent ratios and the residuals become jagged (adjusted $R^2 = 0.20$ for depth pruning in Table~\ref{tab:method_level}). A smoother importance criterion appears to recover the fit -- ShortGPT reaches an average-task adjusted $R^2$ of $0.65$ on GPT-OSS-20B -- so the law should be fitted per operator, or per layer-block segment when the operator is known to be discontinuous.
    \item \textbf{Collapse regions at high pruning ratios.} Beyond $r \approx 70$--$80\%$ performance approaches the random baseline on most tasks, and a power law cannot describe a floor. Predictions in this regime should be treated as extrapolations only; in practice we recommend restricting the fit to ratios below the observed collapse onset.
    \item \textbf{Cliff-like task degradation.} Tasks that stay flat and then collapse, such as CoQA, are fitted with an inflated intercept ($P_0 \gg 1$) and a large exponent. This is a signature of a phase transition rather than of super-compressibility, and pooling such tasks with gradually decaying ones (e.g., CommonsenseQA, $\alpha = 0.52$) distorts both coefficients. Task-level coefficients should be reported separately when the task family is heterogeneous.
    \item \textbf{Calibration for out-of-distribution methods.} Zero-shot transfer to an unseen pruner carries higher error (up to $0.13$ in Table~\ref{tab:extrapolate_method}). One-shot re-estimation of $P_0$ narrows this gap for most methods but not all -- it worsens SVD-LLM ($0.13 \rightarrow 0.16$) -- so it helps only when the calibration ratio is representative of the deployment regime. Re-estimating $\alpha$ as well requires at least two evaluations at well-separated ratios, and is worthwhile only when that budget is available.
\end{enumerate}

A further scope restriction concerns recovery fine-tuning. We fine-tune on WikiText-2 only, and report the resulting coefficients (Appendix~\ref{appx:rft_results}) as an illustration that recovery shifts $\alpha$ and $P_0$ in interpretable ways rather than as a theory of recovery. The scaling behavior of post-pruning recovery as a function of fine-tuning data and model size is treated directly by \citet{chen2024scaling}, whose results are complementary to ours: our law characterizes the pre-recovery starting point, theirs the recovery trajectory.}

\section*{Acknowledgment}
We acknowledge the support of the IBM-IITD AI
Horizons network. T. Chakraborty would like to acknowledge the support of the Yardi School of Artificial Intelligence Travel Grant. 

\bibliographystyle{IEEEtranN}
\small
\bibliography{refs}

\beginappendix
\setcounter{tocdepth}{3}

\section{Dataset Descriptions}
\label{appx:dataset}

The WikiText corpus \citep{merity2016pointer} is a standard benchmark for language modeling, consisting of human-curated articles that are stylistically polished, factually reliable, and written from a neutral perspective. Two variants exist -- WikiText-2 and WikiText-103,with our experiments making use of WikiText-2. The LAMBADA dataset (\textit{Language Modeling Broadened to Account for Discourse Aspects}) \citep{paperno2016lambada} contains narrative passages designed to test a model's ability to predict the final word, a task that requires integrating broad context and maintaining discourse-level coherence. PiQA \citep{bisk2020piqa} evaluates physical common-sense reasoning in everyday situations, often with unconventional but practical solutions. Each instance poses an instruction-based problem (e.g., how to construct, bake, or manipulate objects) and provides two candidate solutions, of which exactly one is correct, framing the task as multiple-choice QA. WinoGrande \citep{sakaguchi2021winogrande}, derived from the Winograd Schema Challenge \citep{levesque2012winograd}, is a large-scale benchmark for pronoun resolution, a task trivial for humans yet challenging for AI systems. HellaSwag \citep{zellers2019hellaswag} focuses on common-sense natural language inference: given a context, models must select the most plausible continuation. The AI2 Reasoning Challenge \citep{clark2018think} provides grade-school science exam questions that require both knowledge and reasoning to solve. Finally, CoQA \citep{reddy2019coqa} is a conversational QA dataset containing 127k question-answer pairs across 8k dialogues spanning seven domains. Its questions are conversational in style, and answers are free-form text supported by highlighted evidence in the passage, making it a benchmark for building dialogue-based reading comprehension systems.

\section{Results}
\label{appx:results}



In this section, we present the detailed task-level results for the pruned LLMs. These results complement the main paper by offering a fine-grained view of performance degradation under pruning, highlighting differences across architectures, pruning ratios, and downstream tasks.

\subsection{Pruned LLM Performance}

\begin{figure*}[!htb]
    \centering
    \subfloat[Reasoning tasks]{\includegraphics[width=0.6\linewidth]{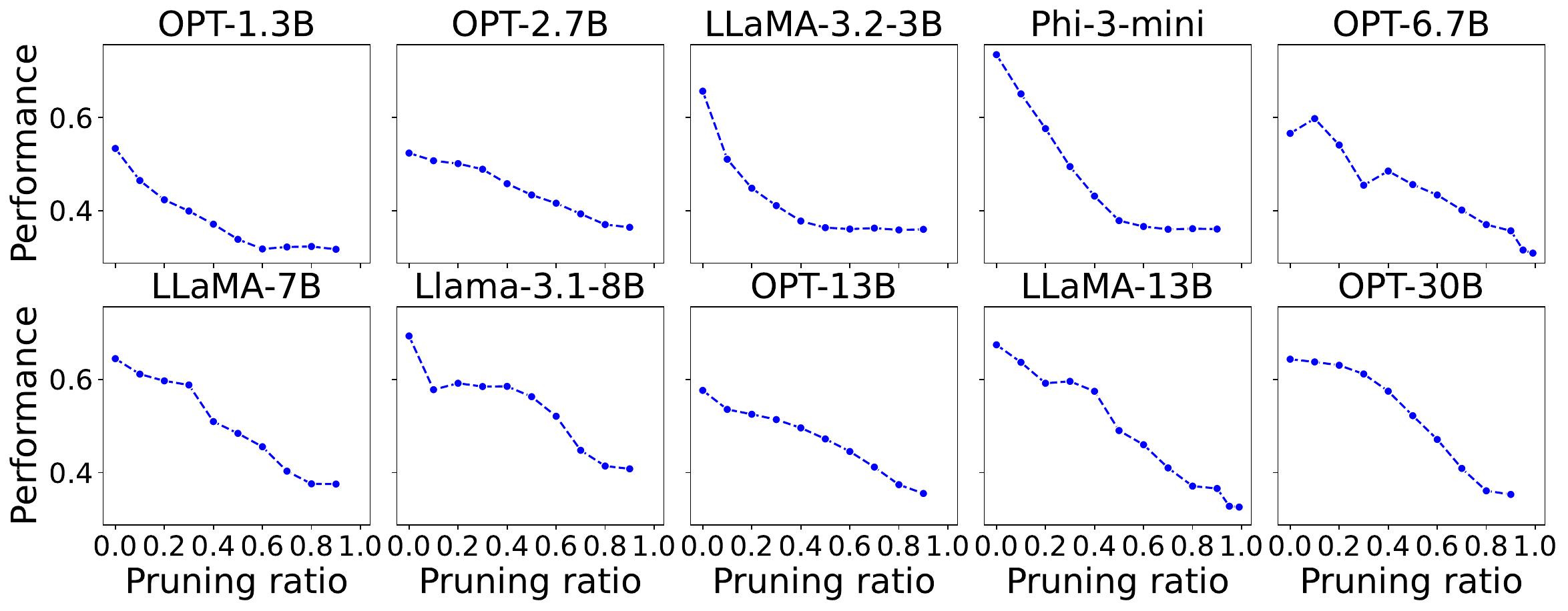}}
    \quad
    \subfloat[QA tasks]{\includegraphics[width=0.6\linewidth]{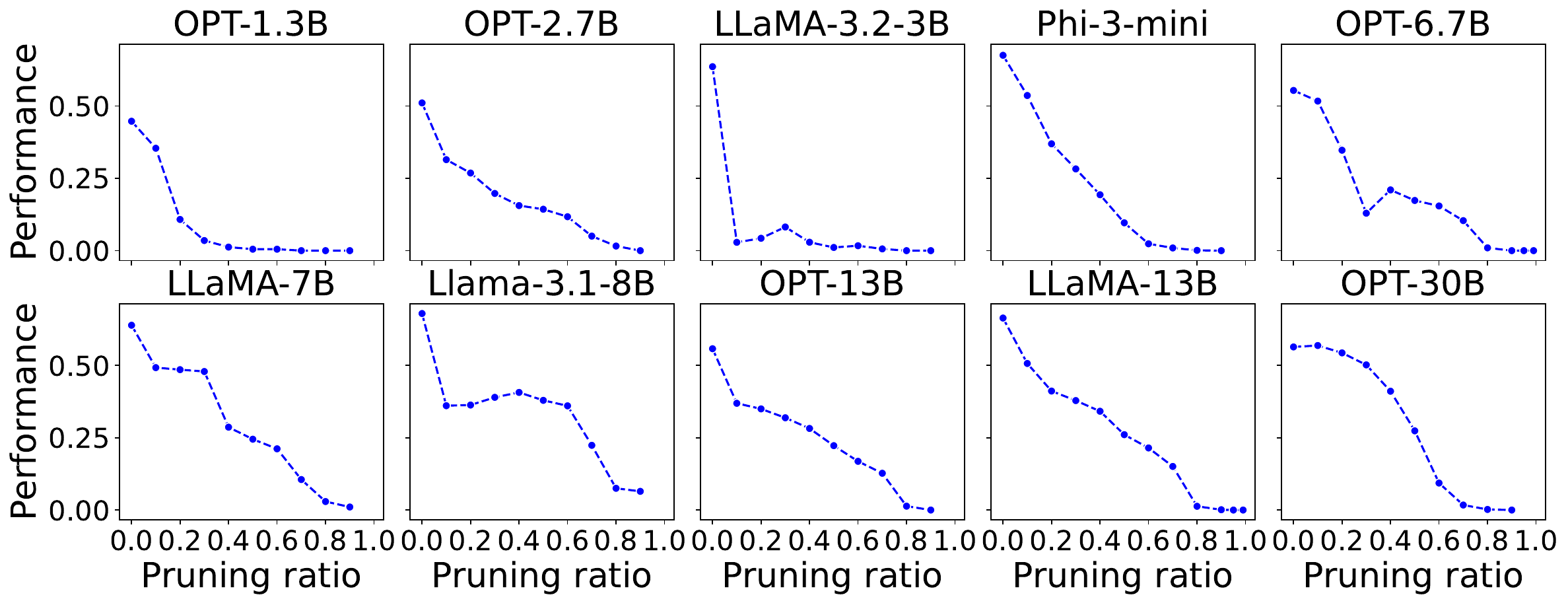}}
    \quad
    \subfloat[Language modeling tasks]{\includegraphics[width=0.6\linewidth]{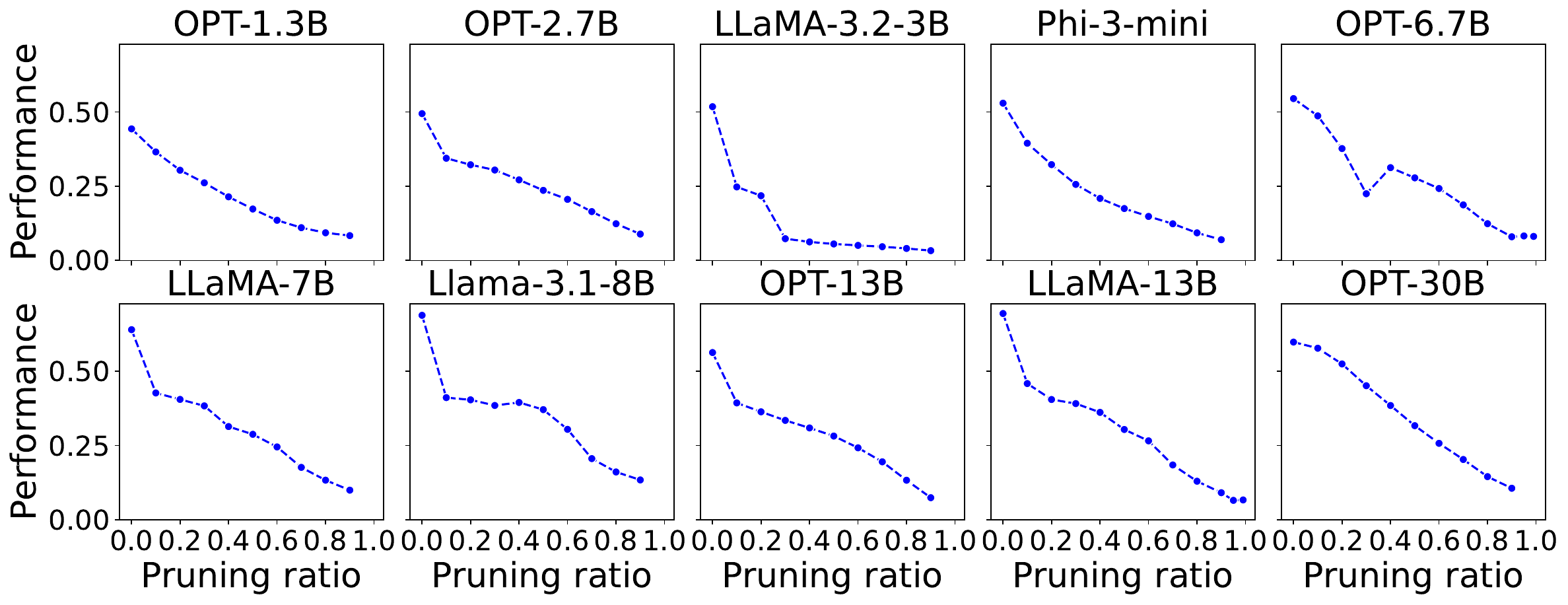}}
    \quad
    \subfloat[Average performance]{\includegraphics[width=0.6\linewidth]{figures/all_model_performance_Average.pdf}}
    \caption{Average downstream performance of different LLMs at varying pruning ratios for all tasks.}
    \label{fig:model_wise_avg_performance_full}
\end{figure*}

Figure~\ref{fig:model_wise_avg_performance_full} provides a comprehensive visual breakdown of downstream performance for all ten evaluated LLMs across varying pruning ratios. 

Reasoning tasks exhibit the most graceful degradation. Larger models like \textbf{OPT-30B} and \textbf{LLaMA-13B} maintain a remarkably flat performance trajectory up to 50-60\% pruning before showing significant decline. Even smaller, highly efficient models like \textbf{Phi-3-mini} and \textbf{LLaMA-3.2-3B} start with high baselines ($>0.6$) and decay relatively smoothly. This visual trend corresponds directly to the low decay exponents ($\alpha \approx 0.22$) reported in the main text, suggesting that logical reasoning relies on distributed representations that are robust to sparse removal.

In sharp contrast, the post-pruning results reveals a ``cliff-like'' failure mode for Question Answering. Almost every model, regardless of scale, suffers a catastrophic collapse once pruning ratios exceed 40-50\%. For instance, \textbf{OPT-1.3B} and \textbf{LLaMA-3.2-3B} see their accuracy plummet to near-random levels almost immediately after the 20\% pruning mark. This brittle behavior visually justifies the high bias terms ($P_0 > 1$) and steep decay coefficients ($\alpha > 2.0$) found in our scaling laws, indicating that the knowledge retrieval circuits required for QA lack the redundancy found in reasoning pathways.

Language modeling occupies the middle ground. The curves are steeper than Reasoning but lack the sudden drop-offs of QA. We observe a quasi-linear decay in log-perplexity (mapped to normalized accuracy here) for models like \textbf{LLaMA-7B} and \textbf{LLaMA-3.1-8B}. This suggests that while linguistic coherence degrades predictably with parameter removal, it does not vanish as abruptly as factual recall.

Comparing the degradation curves across columns highlights the interaction between model scale and architectural efficiency:

\begin{itemize}[leftmargin=*] \setlength{\itemsep}{0pt}
    \item \textbf{Redundancy in Scale:} Larger models (rightmost columns, e.g., OPT-30B) exhibit wider ``safe zones'' -- plateaus where performance remains unchanged despite pruning. This empirically confirms the ``Lottery Ticket'' hypothesis~\citep{frankle2018lottery} at scale, where massive models possess significant sparse subnetworks capable of solving the task.
    \item \textbf{Fragility of Efficient Models:} Newer, parameter-efficient models like \textbf{LLaMA-3.2-3B} and \textbf{Phi-3-mini} show high initial performance but steeper initial slopes compared to older, larger models like OPT-13B. This supports our finding that highly optimized architectures have less ``slack'' or redundancy; they are efficient by design, leaving less room for pruning without immediate penalties.
\end{itemize}

\subsection{Pruning Laws with Recovery Fine-Tuning}
\label{appx:rft_results}

While our main analysis focuses on the direct impact of pruning (zero-shot), practitioners often apply recovery fine-tuning (RFT) to restore lost capabilities. Table~\ref{tab:all_scaling_rft} reports the pruning law coefficients fitted on models after they have undergone recovery fine-tuning on the WikiText-2 dataset. 

\paragraph{Task-level robustness shifts.}
Comparing these RFT coefficients with the zero-shot coefficients in the main text (Table~\ref{tab:task_level}), we observe distinct shifts in task sensitivity:
\begin{itemize}[leftmargin=*] \setlength{\itemsep}{0pt}
    \item \textbf{Language Modeling gains the most resilience.} The decay exponent $\alpha$ for language tasks drops significantly from $\approx 0.73$ (zero-shot) to $\approx 0.47$ (post-RFT). This indicates that fine-tuning is highly effective at repairing the specific probabilistic distributions required for next-token prediction, effectively "flattening" the degradation curve.
    \item \textbf{Hierarchy of sensitivity remains invariant.} Despite the improvements, the relative ordering of task fragility remains unchanged. QA tasks remain the most brittle ($\alpha \approx 2.39$, comparable to the zero-shot $\alpha \approx 2.42$), while Reasoning remains the most robust ($\alpha \approx 0.24$, similar to zero-shot $\alpha \approx 0.22$). This suggests that RFT helps recover general linguistic patterns (Language modeling) but struggles to recover the specific, "cliff-like" knowledge retrieval mechanisms required for QA once they are pruned.
\end{itemize}

\paragraph{Method-level consistency.}
At the method level (Table~\ref{tab:all_scaling_rft}c), Unstructured Pruning continues to exhibit the best fit ($Adj~R^2 \approx 0.88$), though its $\alpha$ coefficient decreases, reflecting improved retention. Interestingly, Depth Pruning sees a slight improvement in goodness-of-fit ($Adj~R^2$ rises from $0.20 \to 0.29$), suggesting that fine-tuning smooths out some of the "step-wise" variance caused by removing discrete layers, though it remains the noisiest method to predict.

\begin{table*} 
\centering
\subfloat[Task-level]{
\scalebox{1}{
    \begin{tabular}{lccrrr}
    \toprule
    \bf Task  & $\alpha$ & $P_0$ & \bf Adj $R^2$ & \bf F Statistic & \bf Test Error \\
    \midrule
    \bf QA    & 2.39 $\pm$ 0.11 & 1.32 $\pm$ 0.15 & 0.91  & 350.23 & 0.03 \\
    \bf Reasoning & 0.24 $\pm$ 0.02 & 0.9 $\pm$ 0.02 & 0.84  & 231.12 & 0.05 \\
    \bf Language & 0.47 $\pm$ 0.01 & 0.86 $\pm$ 0.02 & 0.96  & 1129.47 & 0.01 \\
    \bf Average & 0.27 $\pm$ 0.01 & 0.89 $\pm$ 0.02 & 0.89  & 351.41 & 0.03 \\
    \bottomrule
    \end{tabular}%
 }
}
\quad 
\subfloat[Model-task level]{
\scalebox{1}{
    \begin{tabular}{lllrrr}
      \toprule
    \bf Model & \bf $\alpha$ & $P_0$ & \bf Adj $R^2$ & \bf F Statistic & \bf Test Error \\
    \midrule
    OPT-1.3B & 0.25 $\pm$ 0.04 & 0.65 $\pm$ 0.03 & 0.65  & 24.19 & 0.06 \\
    OPT-2.7B & 0.21 $\pm$ 0.01 & 0.89 $\pm$ 0.02 & 0.96  & 209.1 & 0.02 \\
    LLaMA-3.2-3B & 0.21 $\pm$ 0.04 & 0.51 $\pm$ 0.02 & 0.61  & 37.03 & 0.04 \\
    Phi-3-mini & 0.37 $\pm$ 0.06 & 0.69 $\pm$ 0.03 & 0.71  & 49.31 & 0.05 \\
    OPT-6.7B & 0.28 $\pm$ 0.02 & 0.93 $\pm$ 0.03 & 0.95  & 139.87 & 0.05 \\
    LLaMA-7B & 0.29 $\pm$ 0.03 & 0.87 $\pm$ 0.03 & 0.92  & 98.4 & 0.04 \\
    LLaMA-3.1-8B & 0.35 $\pm$ 0.03 & 0.83 $\pm$ 0.04 & 0.89  & 71.08 & 0.05 \\
    OPT-13B & 0.24 $\pm$ 0.01 & 0.88 $\pm$ 0.01 & 0.99  & 1289.02 & 0.01 \\
    LLaMA-13B & 0.32 $\pm$ 0.03 & 0.88 $\pm$ 0.03 & 0.94   & 133.19 & 0.01 \\
    OPT-30B & 0.52 $\pm$ 0.04 & 0.99 $\pm$ 0.02 & 0.94   & 92.11 & 0.05 \\
    
    \bottomrule
    \end{tabular}%
 }
}
\quad
\subfloat[Method-task level]{
\scalebox{1}{
    \begin{tabular}{lllrrr}
    \toprule
    \textbf{Method} & \textbf{$\alpha$} & $P_0$ & \multicolumn{1}{l}{\textbf{Adj $R^2$}} & \multicolumn{1}{l}{\textbf{F Statistic}} & \multicolumn{1}{l}{\textbf{Test Error}} \\
    \midrule
    Unstructured & 0.34 $\pm$ 0.02 & 1.13 $\pm$ 0.02 & 0.88  & 309.55 & 0.10 \\
     DepthPruning & 0.13 $\pm$ 0.03 & 0.65 $\pm$ 0.03 & 0.29  & 18.61 & 0.06 \\
    WidthPruning & 0.31 $\pm$ 0.03 & 0.88 $\pm$ 0.03 & 0.76  & 141.88 & 0.06\\
    \bottomrule
    \end{tabular}%
    }
}
\caption{Pruning laws of LLMs post-recovery-fine-tuning. All the models are fine-tuned on the WikiText2 dataset with a subsample size of 1000 and a LoRA~\citep{hu2022lora} rank of 8.}
\label{tab:all_scaling_rft}
\end{table*}

\begin{figure*}[t]
    \centering
    \subfloat[Reasoning tasks]{\includegraphics[width=0.6\linewidth]{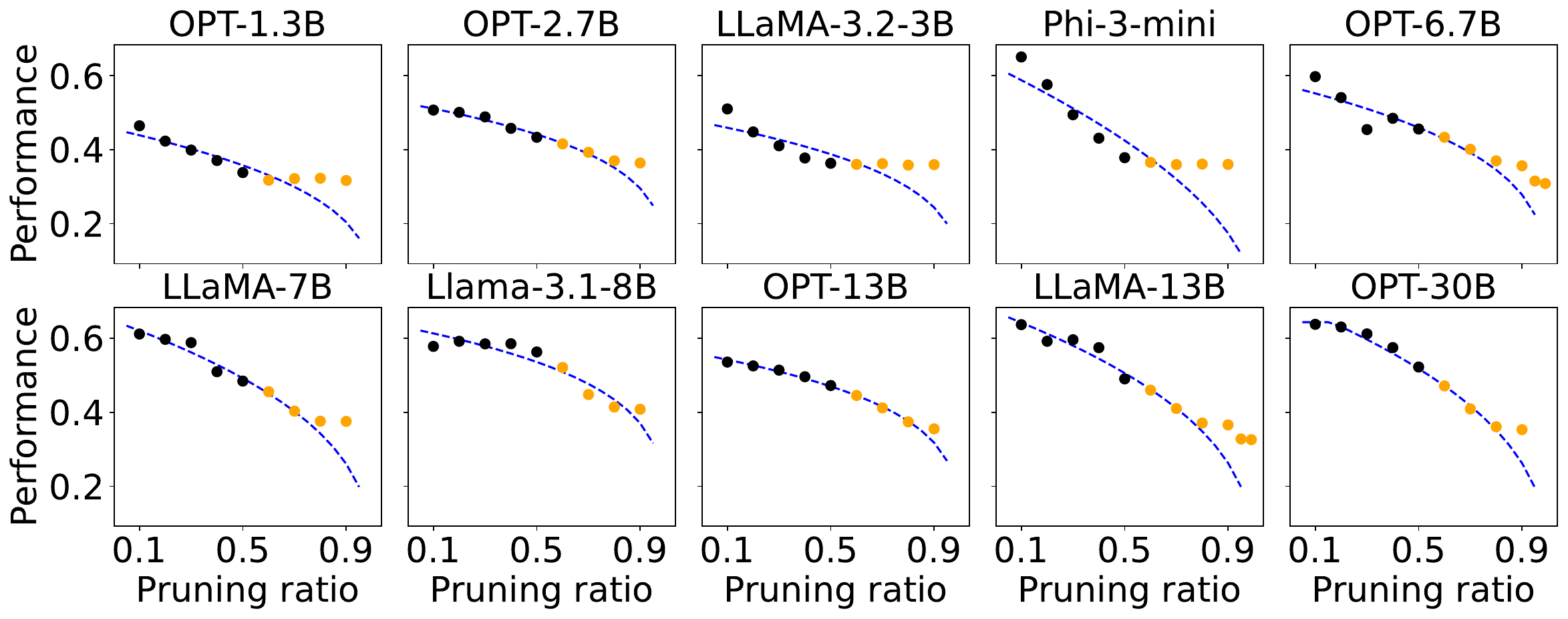}}
    \quad
    \subfloat[QA tasks]{\includegraphics[width=0.6\linewidth]{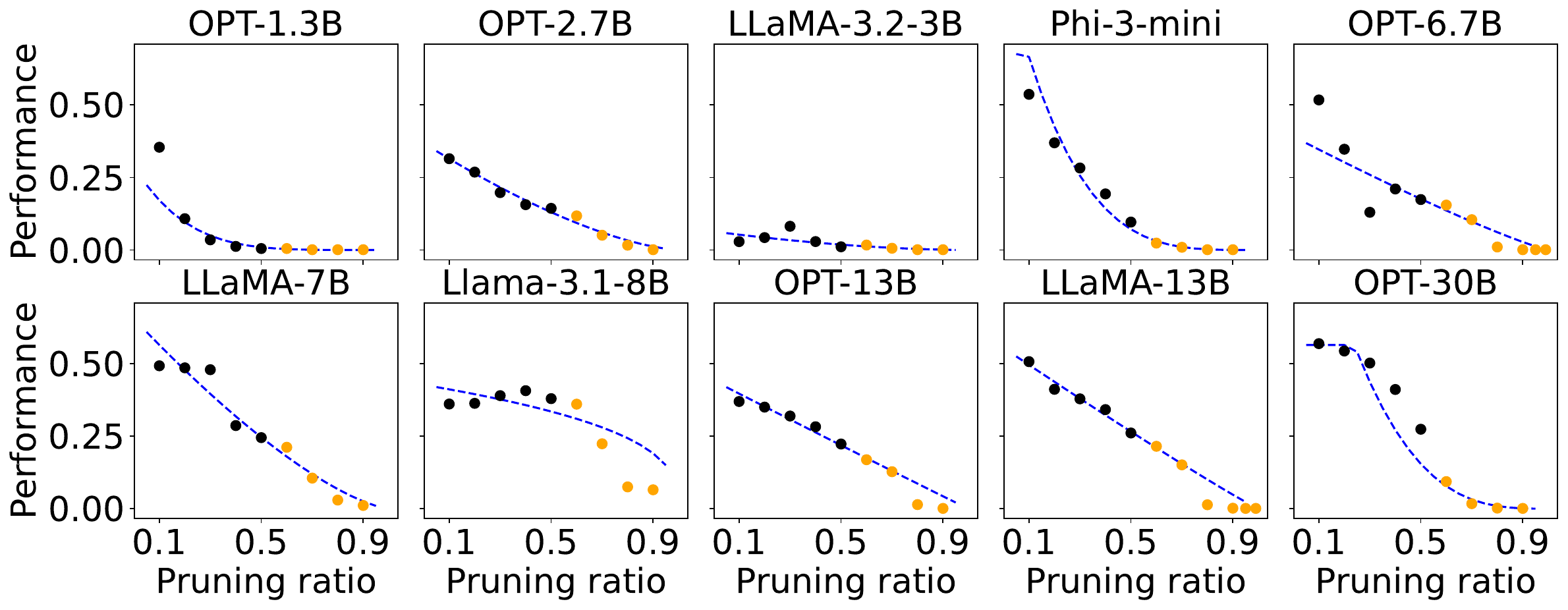}}
    \quad
    \subfloat[Language modeling tasks]{\includegraphics[width=0.6\linewidth]{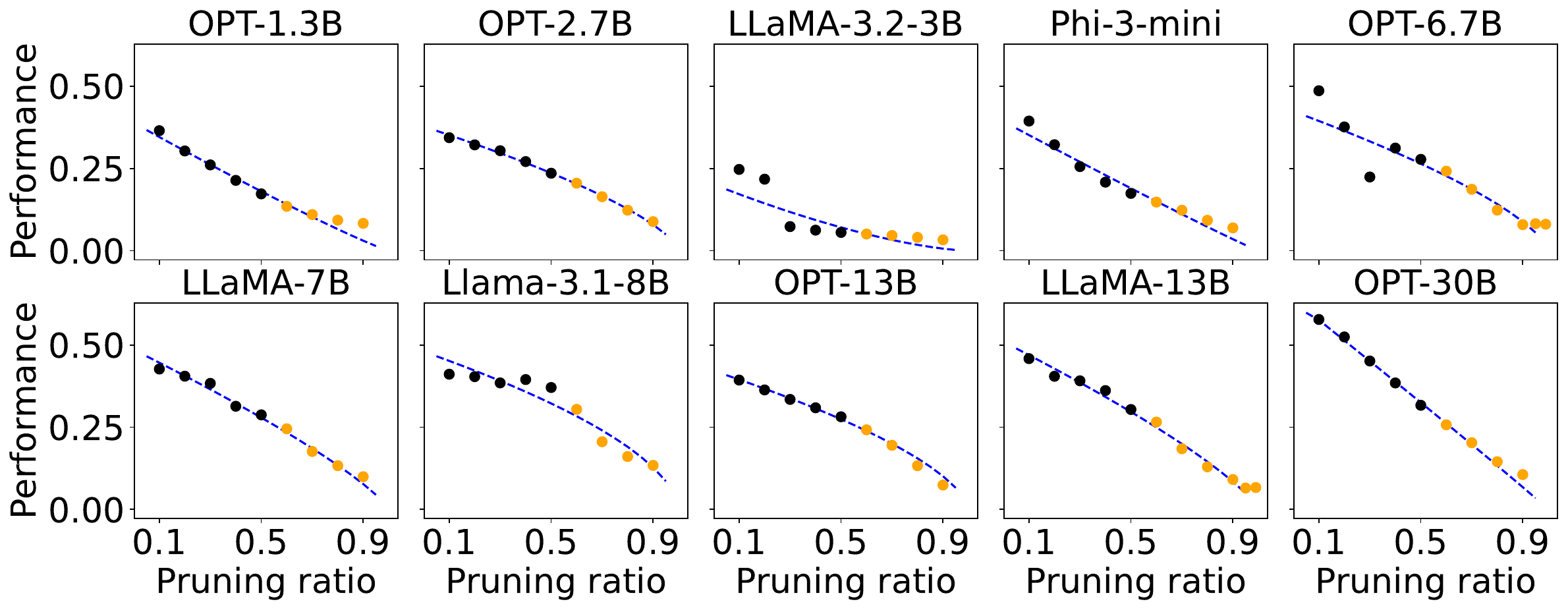}}
    \quad
    \subfloat[Average task performance]{\includegraphics[width=0.6\linewidth]{figures/modelwise_new_Average_scaling.pdf}}
    \caption{Fitted pruning laws for downstream performance of pruned LLMs. \textbf{Black} and \color{orange}orange \color{black}points highlight the training and testing data points, respectively, with \color{black}black \color{black} line indicating the scores predicted by our pruning laws.}
    \label{fig:model_wise_avg_scaling_full}
\end{figure*}

\begin{table*}[!htb] 
  \centering
  \scalebox{1}{
    \begin{tabular}{lcllrrr}
    \toprule
    \textbf{Model} & \multicolumn{1}{l}{\textbf{Method}} & \textbf{$\alpha$} & \textbf{$P_0$} & \multicolumn{1}{l}{\textbf{Adj $R^2$}} & \multicolumn{1}{l}{\textbf{F Statistic}} & \multicolumn{1}{l}{\textbf{Test Error}} \\
    \midrule
    OPT-1.3B & \multirow{10}[0]{*}{Depth} & 0.15 $\pm$ 0.05 & 0.72 $\pm$ 0.05 & 0.52  & 9.62  & 0.28 \\
    OPT-2.7B &       & 0.14 $\pm$ 0.05 & 0.73 $\pm$ 0.05 & 0.51  & 9.27  & 0.2 \\
    LLaMA-3.2-3B &       & 0.14 $\pm$ 0.05 & 0.73 $\pm$ 0.05 & 0.5   & 8.97  & 0.31 \\
    Phi-3-mini &       & 0.14 $\pm$ 0.05 & 0.73 $\pm$ 0.05 & 0.51  & 9.27  & 0.31 \\
    OPT-6.7B &       & 0.72 $\pm$ 0.03 & 0.96 $\pm$ 0.03 & 0.99  & 677.7 & 0.03 \\
    LLaMA-7B &       & 0.17 $\pm$ 0.1 & 0.81 $\pm$ 0.11 & 0.21  & 3.14  & 0.26 \\
    LLaMA-3.1-8B &       & 0.17 $\pm$ 0.1 & 0.81 $\pm$ 0.11 & 0.21  & 3.16  & 0.34 \\
    OPT-13B &       & 0.72 $\pm$ 0.03 & 0.94 $\pm$ 0.03 & 0.99  & 589.92 & 0.04 \\
    LLaMA-13B &       & 0.1 $\pm$ 0.04 & 0.64 $\pm$ 0.04 & 0.48  & 8.35  & 0.15 \\
    OPT-30B &       & 0.73 $\pm$ 0.03 & 0.94 $\pm$ 0.04 & 0.98  & 478.49 & 0.04 \\
    \cdashline{1-7}
    OPT-1.3B & \multirow{10}[0]{*}{Unstr} & 0.01 $\pm$ 0.0 & 0.78 $\pm$ 0.0 & 0.33  & 4.9   & 0.02 \\
    OPT-2.7B &       & 0.01 $\pm$ 0.0 & 0.78 $\pm$ 0.0 & 0.5   & 9.15  & 0.01 \\
    LLaMA-3.2-3B &       & 0.01 $\pm$ 0.0 & 0.78 $\pm$ 0.0 & 0.5   & 9.15  & 0.02 \\
    Phi-3-mini &       & 0.01 $\pm$ 0.0 & 0.78 $\pm$ 0.0 & 0.31  & 4.58  & 0.02 \\
    OPT-6.7B &       & 0.01 $\pm$ 0.0 & 0.84 $\pm$ 0.0 & 0.96  & 176.74 & 0 \\
    LLaMA-7B &       & 0.01 $\pm$ 0.0 & 0.81 $\pm$ 0.0 & 0.95  & 167.97 & 0 \\
    LLaMA-3.1-8B &       & 0.01 $\pm$ 0.0 & 0.81 $\pm$ 0.0 & 0.95  & 167.97 & 0 \\
    OPT-13B &       & 0.01 $\pm$ 0.0 & 0.82 $\pm$ 0.0 & 0.94  & 118.19 & 0.01 \\
    LLaMA-13B &       & 0.02 $\pm$ 0.0 & 0.87 $\pm$ 0.0 & 0.97  & 294.22 & 0 \\
    OPT-30B &       & 0.01 $\pm$ 0.0 & 0.82 $\pm$ 0.0 & 0.94  & 118.19 & 0.01 \\
    \cdashline{1-7}
    OPT-1.3B & \multirow{10}[0]{*}{Width} & -0.0 $\pm$ 0.02 & 1.15 $\pm$ 0.02 & -0.14 & 0.01  & 0.36 \\
    OPT-2.7B &       & -0.0 $\pm$ 0.02 & 1.15 $\pm$ 0.02 & -0.14 & 0.01  & 0.08 \\
    LLaMA-3.2-3B &       & -0.0 $\pm$ 0.02 & 1.14 $\pm$ 0.02 & -0.14 & 0.01  & 0.36 \\
    Phi-3-mini &       & -0.0 $\pm$ 0.02 & 1.15 $\pm$ 0.02 & -0.14 & 0.01  & 0.36 \\
    OPT-6.7B &       & 0.08 $\pm$ 0.02 & 1.08 $\pm$ 0.03 & 0.57  & 11.44 & 0.11 \\
    LLaMA-7B &       & 0.02 $\pm$ 0.02 & 1.02 $\pm$ 0.02 & -0.06 & 0.56  & 0.09 \\
    LLaMA-3.1-8B &       & 0.01 $\pm$ 0.02 & 1.01 $\pm$ 0.02 & -0.09 & 0.31  & 0.23 \\
    OPT-13B &       & 0.14 $\pm$ 0.04 & 1.02 $\pm$ 0.05 & 0.55  & 10.58 & 0.19 \\
    LLaMA-13B &       & 0.15 $\pm$ 0.04 & 0.95 $\pm$ 0.04 & 0.69  & 18.91 & 0.18 \\
    OPT-30B &       & 0.14 $\pm$ 0.04 & 1.03 $\pm$ 0.05 & 0.55  & 10.91 & 0.21 \\
    \bottomrule
    \end{tabular}%
}
  \caption{Coefficients of pruning law for inference speedup of the form $\mathcal{L} = \mathcal{L}_{0}P_{0}(1-r)^{\alpha}$.}
  \label{tab:runtime}
\end{table*}

\begin{figure*}[!htb] 
    \centering
    \subfloat[Unstructured Pruning]{\includegraphics[width=0.7\linewidth]{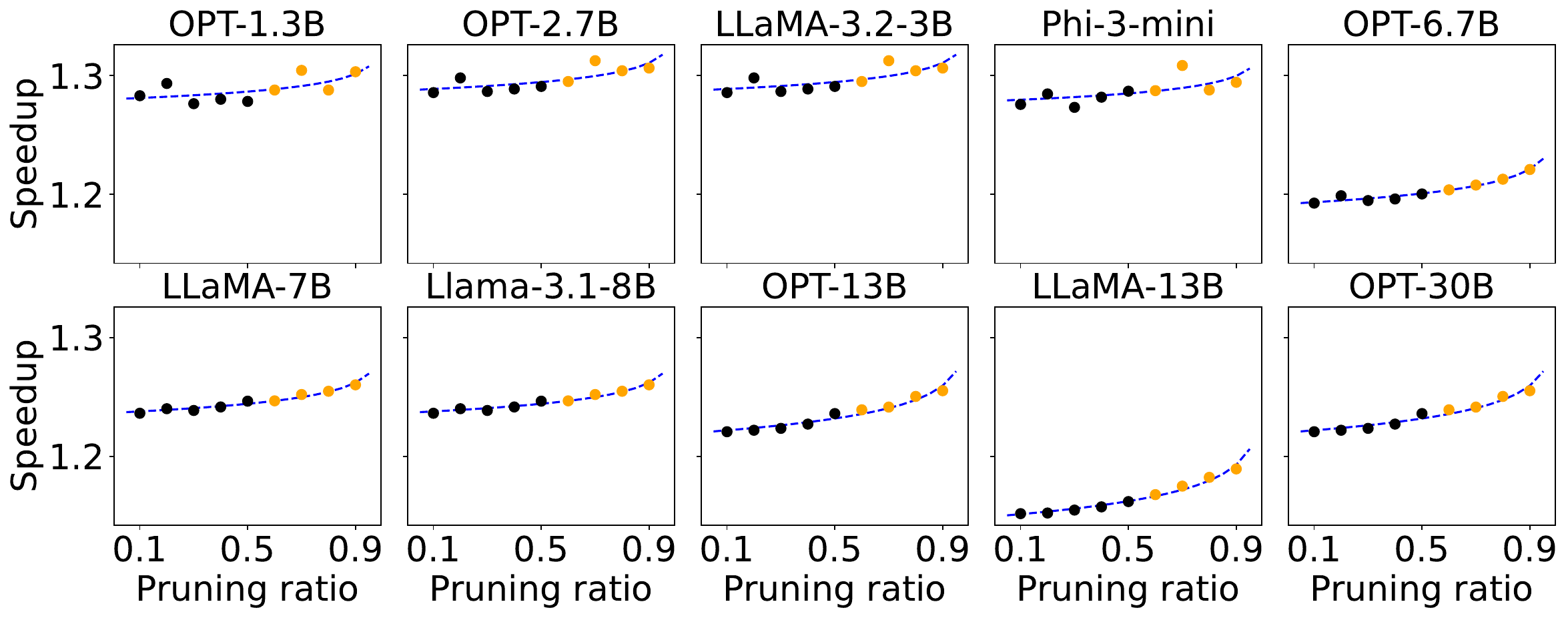}}
    \quad
    \subfloat[Depth Pruning]{\includegraphics[width=0.7\linewidth]{figures/method_modelwise_DepthPruning_Average_scaling_runtime.pdf}}
    \quad
    \subfloat[Width Pruning]{\includegraphics[width=0.7\linewidth]{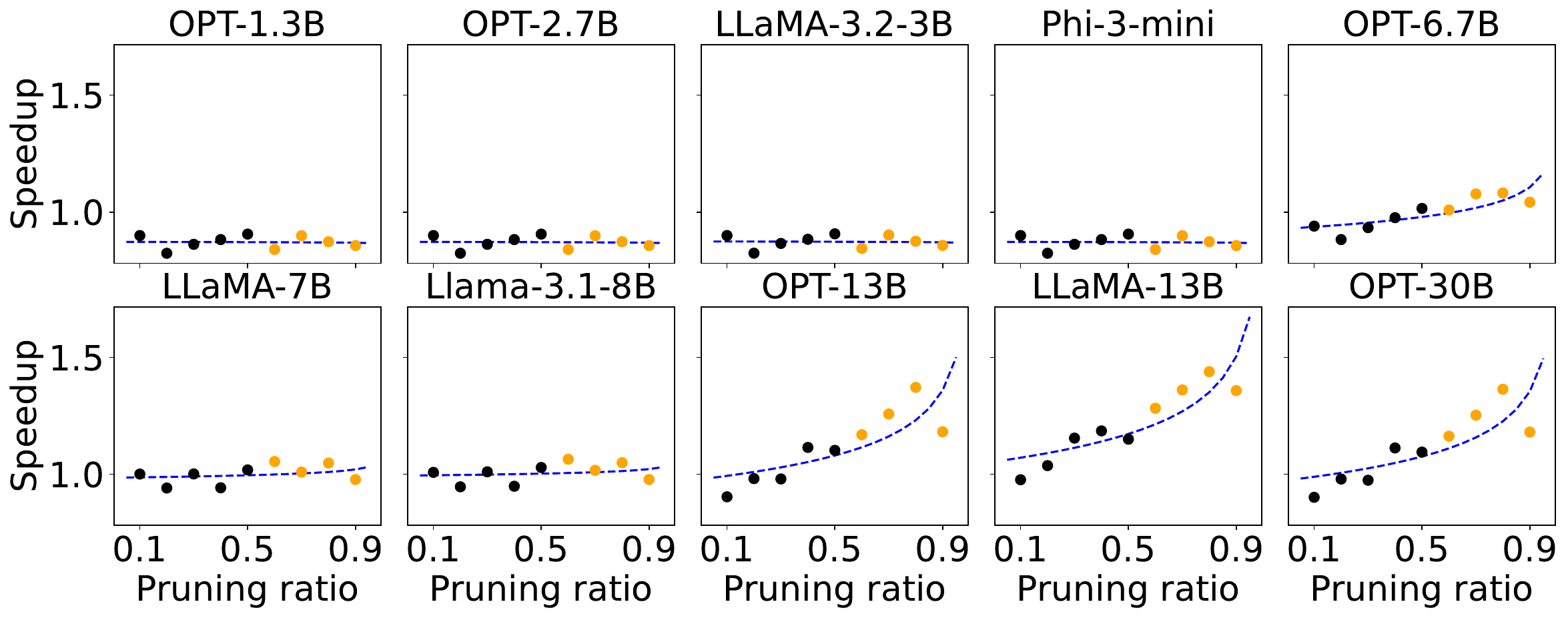}}
    \caption{Fitted pruning laws for inference speedup of pruned LLMs for different pruning methods.}
\label{fig:model_wise_avg_scaling_runtime_full}
\end{figure*}

{\color{black}\subsection{Pruning Laws on a Mixture-of-Experts Model}
\label{appx:moe_results}

Table~\ref{tab:oss_results} reports the coefficients fitted on GPT-OSS-20B, pruned with ShortGPT at ratios $\{10\%,\dots,80\%\}$ and evaluated on seven tasks spanning all three task families. Two observations are worth isolating.

First, the fit is obtained without any change to Equation~\ref{eq:main_fit}. This is not guaranteed a priori: in an MoE model each block routes tokens to a subset of experts, so removing a block both reduces depth and restructures which experts a token can reach, and one might expect the degradation curve to break from the dense case. It does not. The average-task coefficients ($\alpha = 0.48 \pm 0.13$, $P_0 = 0.89 \pm 0.11$) sit inside the range spanned by the dense models in Table~\ref{tab:model_level}, and the task ordering is preserved.

Second, the inclusion of CommonsenseQA separates two degradation styles that were confounded in the main study, where CoQA was our only QA task. CoQA is fitted with $\alpha = 3.53$ and $P_0 = 38.86$, the signature of a curve that stays flat and then collapses; CommonsenseQA is fitted with $\alpha = 0.52$ and $P_0 = 0.54$, a gradual decay closer to the reasoning tasks. Both are described by the same form, but their coefficients differ by an order of magnitude, and the pooled QA group sits between them at an adjusted $R^2$ of $0.55$. This supports reporting task-level rather than family-level coefficients whenever a task family is heterogeneous.

\begin{table}[!htb]
\centering
\color{black}
\subfloat[Task-group level\label{tab:oss_taskgroup}]{
\scalebox{1}{
    \begin{tabular}{lccrrr}
    \toprule
    \bf Task & $\alpha$ & $P_0$ & \bf $R^2$ & \bf F & \bf Error \\
    \midrule
    QA & 0.99 $\pm$ 0.32 & 1.10 $\pm$ 0.27 & 0.55 & 9.7 & 0.09 \\
    Reasoning & 0.30 $\pm$ 0.07 & 0.85 $\pm$ 0.06 & 0.70 & 17.6 & 0.12 \\
    Language & 0.98 $\pm$ 0.14 & 0.79 $\pm$ 0.12 & 0.87 & 48.9 & 0.05 \\
    \midrule
    Average & 0.48 $\pm$ 0.13 & 0.89 $\pm$ 0.11 & 0.65 & 14.0 & 0.13 \\
    \bottomrule
    \end{tabular}%
}
}

\vspace{2mm}

\subfloat[Task level\label{tab:oss_taskwise}]{
\scalebox{1}{
    \begin{tabular}{lccrrr}
    \toprule
    \bf Task & $\alpha$ & $P_0$ & \bf $R^2$ & \bf F & \bf Error \\
    \midrule
    ARC-c & 0.41 $\pm$ 0.17 & 0.69 $\pm$ 0.14 & 0.42 & 6.0 & 0.10 \\
    ARC-e & 0.61 $\pm$ 0.11 & 0.87 $\pm$ 0.09 & 0.82 & 32.8 & 0.11 \\
    PIQA & 0.20 $\pm$ 0.04 & 0.94 $\pm$ 0.03 & 0.81 & 31.5 & 0.10 \\
    WinoGrande & 0.12 $\pm$ 0.04 & 0.86 $\pm$ 0.03 & 0.53 & 8.8 & 0.12 \\
    CoQA & 3.53 $\pm$ 0.71 & 38.86 $\pm$ 1.67 & 0.77 & 24.8 & 0.05 \\
    CSQA & 0.52 $\pm$ 0.24 & 0.54 $\pm$ 0.20 & 0.36 & 4.9 & 0.13 \\
    LAMBADA & 0.98 $\pm$ 0.14 & 0.79 $\pm$ 0.12 & 0.87 & 48.9 & 0.05 \\
    \bottomrule
    \end{tabular}%
}
}
\caption{\color{black}Coefficients of the pruning law fitted on the mixture-of-experts model GPT-OSS-20B~\citep{openai2025gptoss}, pruned with ShortGPT~\citep{men2024shortgptlayerslargelanguage} at ratios $\{10\%,\dots,80\%\}$. The sweep additionally covers CommonsenseQA (CSQA)~\citep{talmor2019commonsenseqa}, widening the QA task family beyond conversational reading comprehension.}
\label{tab:oss_results}
\end{table}

}

\subsection{Justification of the Functional Form}
\label{appx:functional_form}

\begin{table*}[h]
\centering
\scalebox{1}{
\begin{tabular}{lcc}
\toprule
\textbf{Functional Form} & \textbf{Train Error} & \textbf{Test Error} \\
\midrule
$L = L_0 * P_0 * (1 - r)^\alpha$ (Ours) & 0.02 & 0.05 \\
$L = L_0 * (1 - r)^\alpha$ & 0.05 & 0.10 \\
$L = L_0 * (1 - \beta * r^\alpha)$ & 0.03 & 0.05 \\
$L = L_0 * e^{-\beta * r^\alpha}$ & 0.03 & 0.06 \\
$L = L_0 * (1 - \beta * (1 - e^{-\alpha * r}))$ & 0.03 & 0.09 \\
$L = L_0 / (1 + \beta * r^\alpha)$ & 0.03 & 0.07 \\
$L = L_0 / (1 + e^{\beta * (r - \alpha)})$ & 0.03 & 0.06 \\
$L = L_0 * (1 - \alpha * r - \beta * r^2)$ & 0.03 & 0.06 \\
\bottomrule
\end{tabular}
}
\caption{Results with pruning laws of other functional forms. Among all the functional forms investigated, our original functional form generates the lowest training and testing error, affirming the validity of the fitted pruning laws.}
\label{tab:other_forms}
\end{table*}

A core contribution of our work is the specific power-law formulation $\mathcal{L} = \mathcal{L}_0 P_0 (1-r)^\alpha$. To validate this choice, we compared it against seven alternative mathematical models, including standard power laws without bias terms, exponential decays, and rational functions. The results are summarized in Table~\ref{tab:other_forms}.

\paragraph{Necessity of the Bias Term ($P_0$).}
The comparison highlights the critical role of the bias term $P_0$. The standard power law without bias, $\mathcal{L} = \mathcal{L}_0 (1-r)^\alpha$, yields a test error of $0.10$, which is double the error of our proposed form ($0.05$). This confirms our hypothesis that pruning introduces an initial "offset" or phase transition (captured by $P_0$) that a rigid origin-bound power law cannot model accurately.

\paragraph{Superiority over Exponential and Polynomial forms.}
Our formulation also outperforms exponential decay models ($e^{-\beta r^\alpha}$), which tend to under-penalize aggressive pruning, resulting in higher test errors ($0.06-0.09$). Similarly, polynomial expansions ($1 - \alpha r - \beta r^2$) fail to capture the asymptotic tail behavior of degradation as effectively. The proposed form achieves the lowest combined training error ($0.02$) and test error ($0.05$) among all candidates, balancing model complexity with predictive fidelity across the wide range of pruning ratios ($10\%-90\%$).

\section{Practitioner Guidelines for Selecting Pruning Strategies}
\label{appx:guidelines}

To make our findings actionable, we summarize them as practical rules of thumb for real-world adoption of pruning strategies. These guidelines allow practitioners to quickly decide which pruning strategy best matches their constraints and to use the pruning laws for principled planning rather than ad hoc trial-and-error:  

\begin{enumerate}[leftmargin=*] \setlength{\itemsep}{0pt}
    \item \textbf{Selecting the pruning method:}  
    \begin{itemize}[leftmargin=*] \setlength{\itemsep}{0pt}
        \item Use \emph{depth pruning} when runtime speedup is the primary objective, as it yields the largest accelerations (up to $5\times$), though with higher variance in accuracy.  
        \item Use \emph{unstructured pruning} when performance preservation is critical, since it provides the lowest test errors and the most stable predictions, albeit with modest speedups.  
        \item Use \emph{width pruning} for a balanced trade-off between accuracy and speed, offering moderate but consistent improvements.  
    \end{itemize}

    \item \textbf{When facing unseen models:}  
    \begin{itemize}[leftmargin=*] \setlength{\itemsep}{0pt}
        \item Start with \emph{zero-shot mode} by directly reusing fitted coefficients ($\alpha, P_{0}$), which achieves low extrapolation error even across families.  
        \item If limited resources are available to probe the new model, perform \emph{one-shot calibration}: keep $\alpha$ fixed and re-estimate only $P_{0}$ from a single pruning experiment. This is beneficial when accuracy at specific pruning ratios is critical.  
    \end{itemize}

    \item \textbf{When using new pruning methods:}  
    \begin{itemize}[leftmargin=*] \setlength{\itemsep}{0pt}
        \item Begin with zero-shot predictions using coefficients from similar pruning categories.  
        \item Apply one-shot calibration of $P_{0}$ if the method is highly different (e.g., algorithmic pruning heuristics not studied in our experiments) and one evaluation run is feasible.  
    \end{itemize}

    \item \textbf{When targeting different tasks:}  
    \begin{itemize}[leftmargin=*] \setlength{\itemsep}{0pt}
        \item For \emph{reasoning tasks}, higher pruning ratios can be tolerated, as they are more robust (smaller $\alpha$).  
        \item For \emph{QA and language modeling}, adopt more conservative pruning ratios, since these tasks degrade faster with pruning.  
    \end{itemize}

    \item \textbf{Compression planning:}  
    \begin{itemize}[leftmargin=*] \setlength{\itemsep}{0pt}
        \item If multiple evaluations are possible, test at a few pruning ratios (preferably $<50\%$) and fit the pruning law locally to refine estimates.  
        \item If evaluations are infeasible (e.g., resource-constrained scenarios), reuse our published coefficients (model-wise or task-wise) to approximate trade-offs with reasonable accuracy.  
    \end{itemize}
\end{enumerate}

\noindent In summary, practitioners should use zero-shot pruning laws for fast deployment across new models, tasks, or methods, and fall back to one-shot calibration of $P_{0}$ when some evaluation budget is available. This ensures a principled, universal, and resource-aware approach to model pruning.

\end{document}